\documentclass[10pt,journal,compsoc]{IEEEtran}

\usepackage{graphicx}
\usepackage{amsmath}
\usepackage{amssymb}
\usepackage{booktabs} %
\usepackage{pifont}   %
\usepackage{url}
\usepackage{color}
\usepackage{setspace}
\usepackage{multirow}
\usepackage{threeparttable}
\usepackage[ruled]{algorithm2e}
\usepackage[T1]{fontenc}
\usepackage{microtype}
\usepackage{bm}
\usepackage{ulem}
\usepackage[table,xcdraw]{xcolor}
\usepackage{wrapfig}
\usepackage{ragged2e}
\usepackage{adjustbox}
\usepackage{subcaption}

\usepackage{caption}
\ifCLASSOPTIONcompsoc
  \usepackage[nocompress]{cite}
\else
  \usepackage{cite}
\fi

\usepackage[colorlinks,
            linkcolor=red,
            anchorcolor=blue,
            citecolor=teal]{hyperref}

\usepackage[capitalize,noabbrev]{cleveref}

\DeclareMathOperator*{\argmax}{arg\,max}

\begin{document}

\title{DynaPURLS: Dynamic Refinement of Part-aware Representations for Skeleton-based Zero-Shot Action Recognition}

\author{Jingmin Zhu,
        Anqi Zhu,
        James Bailey,
        Jun Liu,
        Hossein Rahmani,
        Mohammed Bennamoun,
        Farid Boussaid,
        and Qiuhong Ke$^*$%
\IEEEcompsocitemizethanks{\IEEEcompsocthanksitem J. Zhu and A. Zhu are with Monash University.\protect\\
E-mail: jingmin.zhu1@monash.edu, maggie.zhu@monash.edu
\IEEEcompsocthanksitem J. Bailey is with Monash University.\protect\\
E-mail: james.a.bailey@monash.edu
\IEEEcompsocthanksitem H. Rahmani and J. Liu are with Lancaster University.\protect\\
E-mail: h.rahmani@lancaster.ac.uk, j.liu81@lancaster.ac.uk
\IEEEcompsocthanksitem M. Bennamoun and F. Boussaid are with University of Western Australia.\protect\\
E-mail: mohammed.bennamoun@uwa.edu.au, farid.boussaid@uwa.edu.au
\IEEEcompsocthanksitem Q. Ke is with Monash University.\protect\\
E-mail: Qiuhong.Ke@monash.edu
\IEEEcompsocthanksitem $^*$ Corresponding author.}%

\IEEEtitleabstractindextext{%
\begin{abstract}
\justifying
Zero-shot skeleton-based action recognition (ZS-SAR) is fundamentally constrained by prevailing approaches that rely on aligning skeleton features with static, class-level semantics. This coarse-grained alignment fails to bridge the domain shift between seen and unseen classes, thereby impeding the effective transfer of fine-grained visual knowledge. To address these limitations, we introduce \textbf{DynaPURLS}, a unified framework that establishes robust, multi-scale visual-semantic correspondences and dynamically refines them at inference time to enhance generalization. Our framework leverages a large language model to generate hierarchical textual descriptions that encompass both global movements and local body-part dynamics. Concurrently, an adaptive partitioning module produces fine-grained visual representations by semantically grouping skeleton joints. To fortify this fine-grained alignment against the train-test domain shift, DynaPURLS incorporates a dynamic refinement module. During inference, this module adapts textual features to the incoming visual stream via a lightweight learnable projection. This refinement process is stabilized by a confidence-aware, class-balanced memory bank, which mitigates error propagation from noisy pseudo-labels. Extensive experiments on three large-scale benchmark datasets, including NTU RGB+D 60/120 and PKU-MMD, demonstrate that DynaPURLS significantly outperforms prior art, setting new state-of-the-art records. The source code is made publicly available at \url{https://github.com/Alchemist0754/DynaPURLS}
\end{abstract}}

}

\maketitle

\IEEEdisplaynontitleabstractindextext
\IEEEpeerreviewmaketitle

\IEEEraisesectionheading{\section{Introduction}\label{sec:introduction}}
\IEEEPARstart{H}{uman} Action Recognition (HAR) has emerged as a cornerstone technology in computer vision, enabling machines to comprehend human behaviors from sensory data. This capability is fundamental to transformative applications across diverse domains, including immersive virtual reality \cite{vr1, vr2}, intelligent transportation \cite{autodrive1, autodrive2}, large-scale video retrieval \cite{retrieval}, and human-robot collaboration \cite{robotics1, robotics2}. While RGB-based solutions have historically dominated HAR research, buoyed by massive annotated datasets \cite{r2+1d, pyramid}, the maturation of human pose estimation \cite{openpose} and depth-sensing hardware \cite{ntu60, ntu120} has positioned 3D skeleton sequences as a compelling alternative. Skeleton data offers distinct advantages: it is computationally efficient, inherently privacy-preserving, and robust to environmental variations such as background clutter and lighting changes \cite{app1, app2, app3}.

\begin{figure}[t]
\centering
\includegraphics[width=0.95\linewidth]{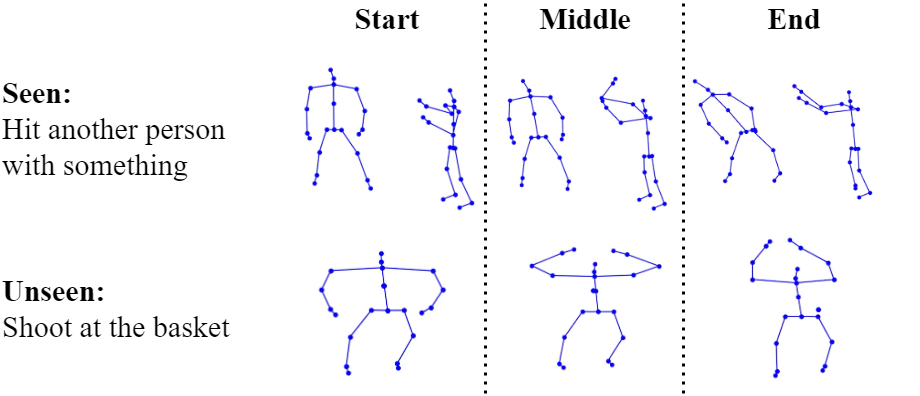}
\caption{An example illustrating the limitations of global-only alignment. A seen class (``Hit another person with something'') and an unseen class (``Shoot at the basket'') from \textit{NTU RGB+D 120} \cite{ntu120} exhibit different overall motions but share similar local hand movements.}
\label{fig:zsl}
\vspace{-0.2cm}
\end{figure}

Despite remarkable progress in fully supervised skeleton-based HAR \cite{deep-skel, deep-skel2, deep-skel3, 2s-gcn, dg-gcn, shift-gcn}, these approaches are constrained by their reliance on exhaustive annotations for all action classes. This paradigm falters when confronted with real-world scenarios involving rare, hazardous, or expensive-to-collect actions, creating a critical deployment bottleneck. Zero-Shot Learning (ZSL) offers a compelling path forward by transferring knowledge from seen to unseen categories through a shared semantic space, enabling the recognition of novel actions without direct training examples.

\begin{figure}[t]
    \centering
    \footnotesize
    \begin{subfigure}[b]{0.44\linewidth}
        \centering
        \includegraphics[width=\textwidth, trim=100 100 100 100, clip]{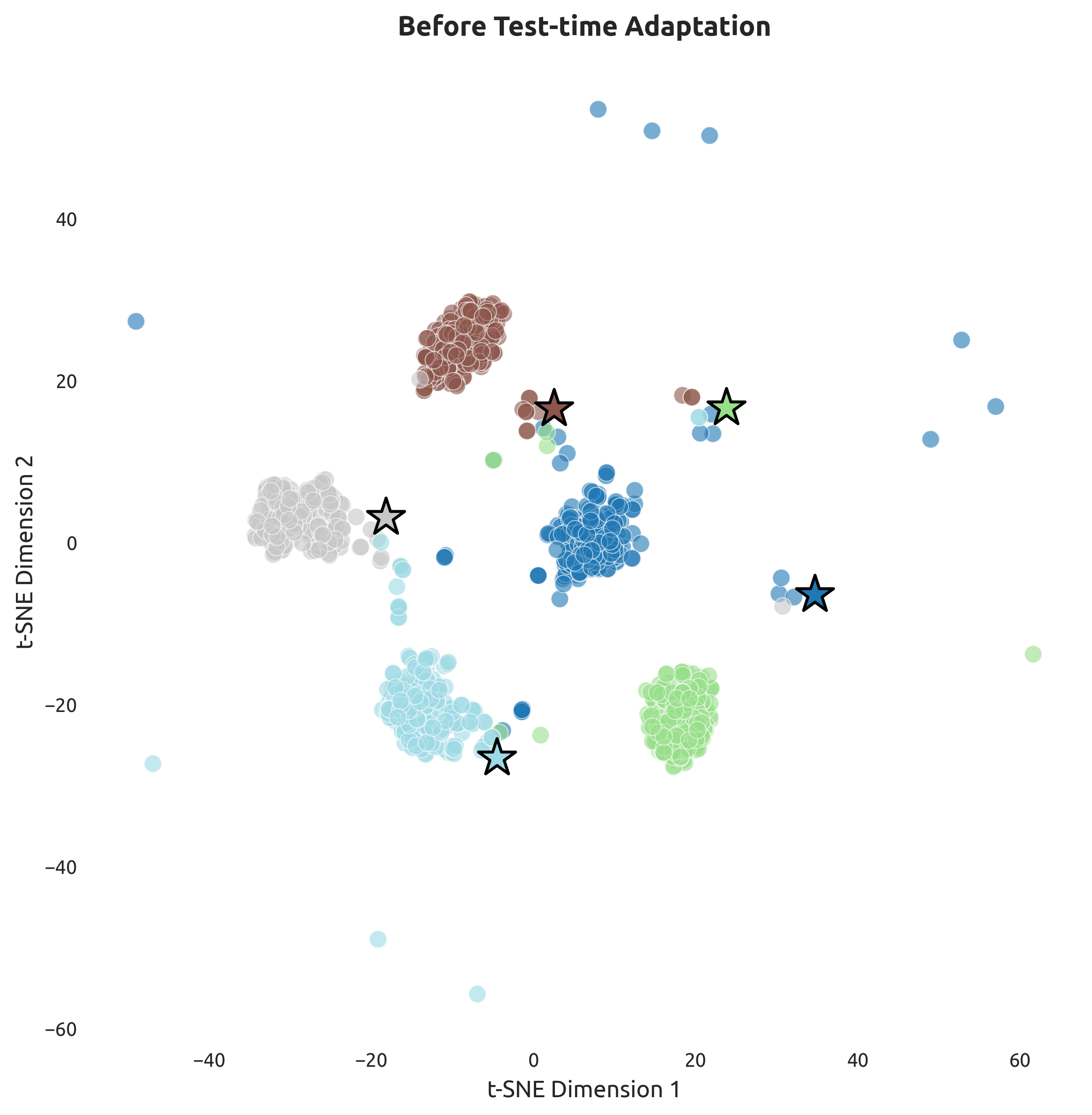}
        \caption{Pre-adaptation}
    \end{subfigure}
    \hfill
    \begin{subfigure}[b]{0.44\linewidth}
        \centering
        \includegraphics[width=\textwidth, trim=100 100 100 100, clip]{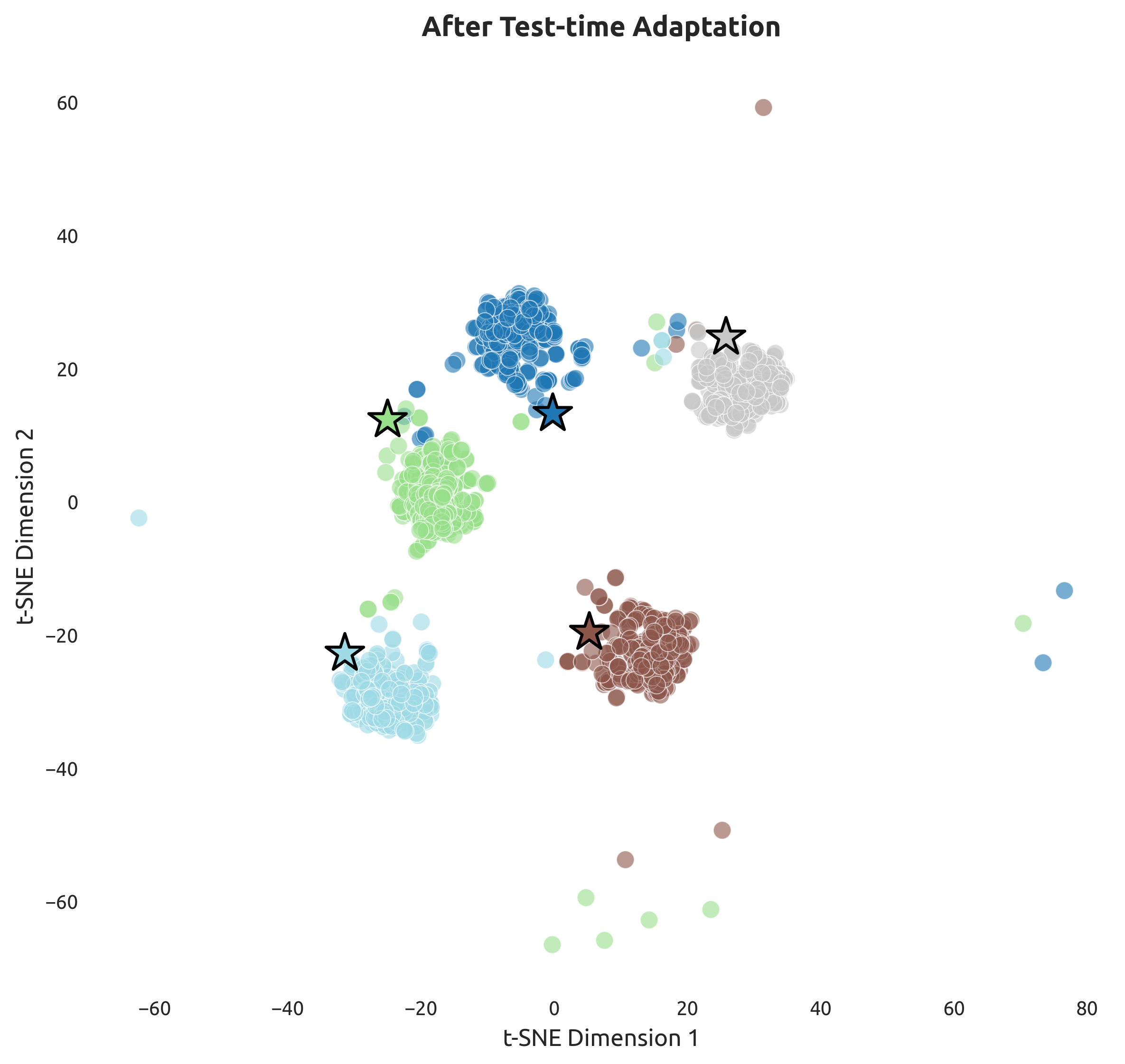}
        \caption{Post-adaptation}
    \end{subfigure}
    \caption{t-SNE visualization of feature distributions before and after test-time refinement on NTU RGB+D 60 (55/5 split), where $\star$ represents text features and $\bullet$ represents skeleton features. Our method effectively promotes alignment between modalities, achieving a 9.3\% improvement.}
    \label{fig:tsne}
    \vspace{-0.2cm}
\end{figure}

However, existing skeleton-based ZSL methods \cite{jpose, synse-zsl, smie} predominantly depend on aligning a single, global motion feature with a corresponding class-level semantic embedding. This coarse alignment strategy is insufficient for the nuanced demands of action recognition. As illustrated in \cref{fig:zsl}, actions with disparate global patterns (e.g., ``Hit another person with something'' vs. ``Shoot at the basket'') may share remarkably similar fine-grained motions in specific body parts. Such part-level correspondences are a crucial source of transferable knowledge, yet they are largely neglected by current models. Moreover, even recent fine-grained ZSL methods \cite{li2024sadvaeimprovingzeroshotskeletonbased,chen2024finegrainedinformationguideddualprompts} introduce another critical challenge by operating on static semantic information. The textual descriptions are generated offline from language prompts alone, without access to the visual data they are meant to represent. This often results in descriptions that are semantically plausible but lack the subtle, instance-specific distinctions necessary for precise recognition. The learned alignment is often brittle and fails to generalize, as the features of unseen actions introduce a significant domain shift, preventing effective knowledge transfer.

To overcome these interconnected challenges, we introduce \textbf{DynaPURLS}, a framework that pioneers a paradigm shift from static matching to dynamic, adaptive alignment. To address the issue of coarse alignment, DynaPURLS forges a more granular connection between semantics and motion. It employs a Large Language Model (LLM), specifically GPT-3 \cite{gpt3}, to decompose high-level action labels into detailed descriptions of both global movements and localized part-level motions. These multi-granularity descriptions, encoded via text encoders like CLIP \cite{clip}, are then aligned with corresponding visual features. Critically, instead of using fixed body parts, DynaPURLS utilizes an adaptive partitioning module that learns to group skeleton joints, creating visual representations that dynamically match the generated semantic descriptions. This ensures a robust, fine-grained alignment during training.

Furthermore, DynaPURLS tackles the problem of static semantics by introducing a \textbf{dynamic, confidence-guided refinement at inference time}. This novel mechanism directly addresses the domain shift problem by introducing a lightweight transformation that dynamically refines the fine-grained textual features to adapt to the specific visual context of each sample. This refinement markedly enhances alignment and modality coherence for unseen classes, as visualized in \cref{fig:tsne}. To ensure a stable optimization process and mitigate bias, our methodology is fundamentally architected on a principle of \textbf{class-balancing}. The core of this is a \textbf{confidence-guided optimization strategy} that leverages the model's own high-confidence predictions as a self-supervised signal. We innovate by incorporating a small-scale, \textbf{class-balanced memory bank} to mechanistically avert the risk of overconfidence. This design filters and stores high-confidence samples, enforcing a balanced representation across classes during updates, which not only prevents model skew but also \textbf{drastically improves the recognition rate of previously misclassified minority classes}.

In summary, our main contributions are:

\begin{itemize}
    \item We propose \textbf{DynaPURLS}, a novel framework for zero-shot skeleton-based action recognition that addresses the limitations of coarse alignment and static semantics by creating a dynamic link between visual and textual features.

    \item We introduce a \textbf{fine-grained adaptive alignment} strategy that matches multi-granularity semantics from an LLM with visual features from a novel \textbf{adaptive partitioning module} to enable effective part-level knowledge transfer.

    \item We pioneer a \textbf{test-time feature refinement} mechanism using a lightweight, \textbf{confidence-guided optimization} and a \textbf{class-balanced memory bank} to adapt semantic features at inference, mitigating domain shift with minimal overhead.

    \item Through extensive experiments on three challenging benchmarks (NTU RGB+D 60 \cite{ntu60}, NTU RGB+D 120 \cite{ntu120}, and PKU-MMD \cite{liu2017pku}), we demonstrate that DynaPURLS sets a new \textbf{state of the art} in both ZSL and GZSL settings.
\end{itemize}

This paper is an extension of our earlier conference paper~\cite{zhu2024partawareunifiedrepresentationlanguage}. The main improvements are: \textbf{(1)} A novel dynamic refinement module for inference time that resolves the static semantic limitations of the original PURLS and yields consistent performance gains. To our knowledge, this is the first work to employ test-time adaptation for zero-shot skeleton-based action recognition. \textbf{(2)} An extended evaluation under the GZSL setting demonstrates that our method achieves state-of-the-art results, outperforming PURLS and prior methods in a more realistic recognition scenario. \textbf{(3)} A comprehensive suite of ablation studies systematically analyzes each component of our dynamic model, elucidating the design rationale and empirically validating their effectiveness.
\section{Related Work}
\label{sec:related_work}

\subsection{Multi-modal Learning and Zero-Shot Foundations}
\label{sec:rw_foundations}

Zero-Shot Learning (ZSL) recognizes novel classes by transferring knowledge from seen classes via a shared semantic space. Early embedding-based methods focused on learning compatible visual-semantic projections~\cite{emb1, emb2, emb3}, while generative approaches synthesize visual features from semantic embeddings~\cite{aezsl}. Recent compositional action recognition work~\cite{yan2023progressive} has demonstrated the value of progressive instance-aware feature learning for novel action-object combinations. The modern ZSL landscape has been reshaped by large-scale multi-modal learning. CLIP~\cite{clip} created a powerful vision-language manifold through contrastive pre-training, spurring efforts to extend this to other modalities like 3D point clouds~\cite{ulip, point-clip}.

Our work builds upon these principles but argues that generic vision-language models are insufficient for skeleton-based action recognition, which requires specialized designs to exploit fine-grained spatio-temporal characteristics.

\subsection{Zero-Shot Learning for Skeleton-Based Actions}
\label{sec:rw_skeleton_zsl}
Adapting ZSL principles to skeleton-based action recognition began with pioneering works that projected globally-pooled skeleton features into a semantic space using DeViSE-like models or common-space metric learning~\cite{jpose, wray2019fine}. While these methods validated the feasibility of skeleton ZSL, they were limited by their reliance on coarse, global representations, often failing to distinguish actions with similar overall movements but different local limb motions. The importance of spatiotemporal modeling in skeleton understanding has been well-established across various tasks, including motion prediction where co-attention mechanisms~\cite{shu2022spatiotemporal} have proven effective at capturing joint-level spatial consistency and temporal evolution. Beyond supervised settings, self-supervised and semi-supervised approaches have explored multi-granularity representation learning for skeleton data. For instance, MAC-Learning~\cite{shu2023multigranularity} introduced anchor-contrastive learning across local, context, and global granularities to address pair ambiguity in semi-supervised settings, while X-CAR~\cite{xu2022xinvariant} proposed learnable augmentation strategies to ensure consistency between augmentation and representation learning. Similarly, SDS-CL~\cite{gao2023spatiotemporal} decoupled spatial and temporal features for more nuanced contrastive learning. These works underscore the importance of fine-grained, multi-level feature extraction, a principle we leverage in our zero-shot framework. Subsequent research has progressively sought more fine-grained alignment. For instance, SMIE~\cite{smie} improved temporal modeling by maximizing the mutual information between visual and textual distributions, while SynSE-ZSL~\cite{synse-zsl} made an early attempt at local semantic matching by leveraging the syntactic structure (verbs and nouns) within action labels to guide knowledge transfer. The field took a significant leap forward with the use of Large Language Models (LLMs) to generate rich, multi-granularity semantic descriptions. Our foundational work, \textbf{PURLS}~\cite{zhu2024partawareunifiedrepresentationlanguage}, and the concurrent STAR~\cite{chen2024finegrainedinformationguideddualprompts}, represent the state of the art by aligning fine-grained, spatio-temporal skeleton features with detailed, LLM-generated prompts, proving the efficacy of multi-level cross-modal alignment.

Despite their increasing sophistication, these frameworks, including our own PURLS, have largely been built upon \textbf{static semantic representations}. Recognizing this limitation, the most recent works have begun to challenge this static paradigm during the model training phase. For example, SCoPLe~\cite{Zhu_2025_CVPR} introduces learnable cross-modal prompts to refine semantic guidance, and Neuron~\cite{chen2024neuronlearningcontextawareevolving} proposes a context-aware evolving representation. While these methods embed adaptability into the training process, the learned semantic anchors remain fixed during inference. This still leaves a critical ``Semantic-Visual Gap'' when encountering the dynamic visual manifestations of unseen classes. Our work addresses this gap from a distinct angle, proposing a method to overcome the static representation problem directly at \textbf{inference-time}, thereby tackling the domain shift inherent in ZSL more effectively.

\subsection{Test-Time Adaptation for Zero-Shot Learning}
\label{sec:rw_tta_zsl}
Test-Time Adaptation (TTA) aims to bridge the distribution gap between training and test data by adapting a pre-trained model during inference~\cite{niu2022efficient,wang2020tent,gong2022note,wang2022continual}. These general-purpose methods were not designed for the unique challenges of skeleton-based ZSL. Many approaches adapt the model using pseudo-labels derived from its own confident predictions, for instance by minimizing prediction entropy or using nearest-neighbor classification~\cite{zhang2023adanpc}. A prominent trend in vision-language models involves adapting textual prompts to better match test data, using augmentation-based consistency~\cite{shu2022test}, diffusion model-based augmentations~\cite{feng2023diversedataaugmentationdiffusions}, or other prompt enhancement strategies~\cite{guo2022calipzeroshotenhancementclip}. Other efficient methods directly refine predictions by aligning test feature distributions or constructing feature caches~\cite{karmanov2024efficient}. However, when applied to skeleton ZSL, these generic strategies fall short. They either incur significant computational overhead or treat features holistically, failing to perform the fine-grained, targeted adjustments needed to bridge the specific ``Semantic-Visual Gap'' between static, multi-level textual prompts and dynamic skeleton motions.

To address this specific challenge, we propose \textbf{DynaPURLS}, a novel TTA strategy deeply integrated with our PURLS backbone. Rather than applying a generic adaptation scheme, DynaPURLS is purpose-built to resolve the core limitation of PURLS: its reliance on static semantic representations. Its core novelty is being the first to propose adapting the \textbf{multi-granularity semantic representations themselves} at test time. This is achieved through a lightweight, gradient-guided online optimization that transforms semantic features to align with incoming visual evidence, avoiding costly updates to the large visual and text encoders. This dynamic refinement is stabilized by a confidence-guided optimization mechanism and a class-balanced memory bank, ensuring robust and efficient adaptation tailored to the structured nature of skeleton data. DynaPURLS thus carves a distinct path by showing that a specialized, semantics-focused adaptation is significantly more effective for skeleton-based ZSL.

\section{Methodology}
\label{sec:methodology}
\begin{figure}[t]
\centering
\adjustbox{width=0.98\columnwidth,trim=0 0 {0.1\width} 0,clip}{%
    \includegraphics{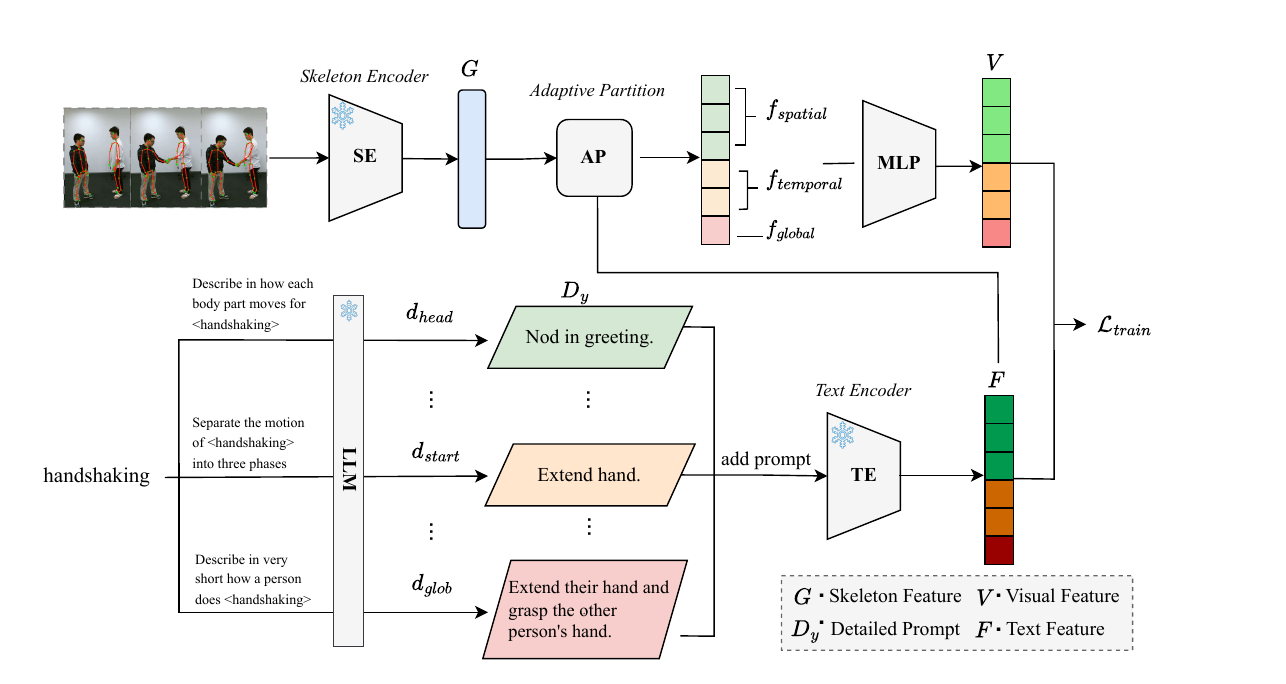}%
}
\caption{\textbf{Training pipeline of DynaPURLS.} The model learns a robust static alignment between visual features extracted by the skeleton encoder and multi-granularity semantic embeddings from the text encoder. We employ an adaptive partitioning module where static text embeddings serve as queries to flexibly aggregate visual information.}
\label{fig:pipeline_training}
\vspace{-0.2cm}
\end{figure}

\begin{figure*}[t]
\centering
\includegraphics[width=\textwidth]{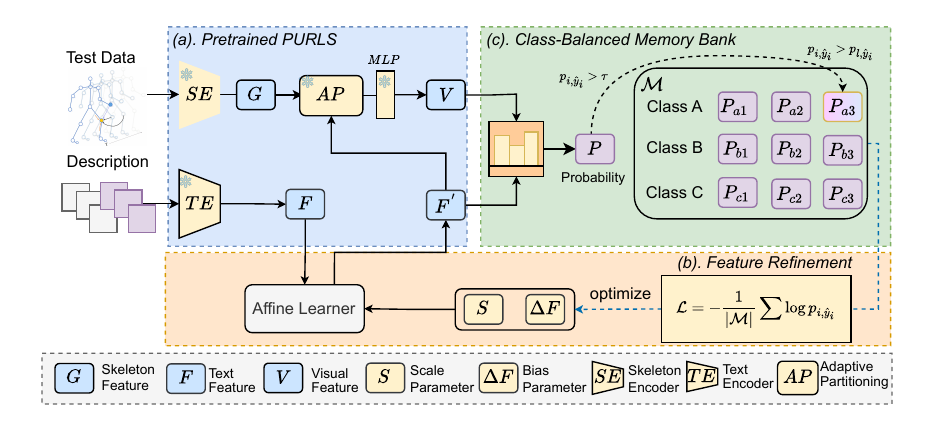}
\caption{
    \textbf{Test-time feature refinement pipeline of DynaPURLS.} 
    The framework dynamically refines semantic embeddings based on high-confidence test samples stored in a class-balanced memory bank, enabling robust adaptation to distribution shifts in unseen classes. The refined embeddings are used to update the cross-modal attention mechanism for improved visual-semantic alignment.
}
\label{fig:pipeline}
\end{figure*}

In this section, we introduce our novel framework, the \textbf{Dyna}mic \textbf{P}art-aware \textbf{U}nified \textbf{R}epresentation between \textbf{L}anguage and \textbf{S}keleton (\textbf{DynaPURLS}), designed for zero-shot skeleton-based action recognition. As shown in Fig.~\ref{fig:pipeline_training} and Fig.~\ref{fig:pipeline}, the framework contains a multi-granularity semantic representation module, which leverages Large Language Models (LLMs) to generate rich, hierarchical action descriptions that capture transferable motion primitives; an adaptive partitioning and visual-semantic alignment module, which employs a cross-modal attention mechanism to dynamically align visual features with their corresponding semantic concepts, overcoming the limitations of static partitioning; and a dynamic test-time feature refinement module, which adapts the semantic embeddings online using a confidence-guided, class-balanced memory bank for robust handling of domain shifts inherent to unseen classes. We first describe the problem definition (Section~\ref{sec:problem_definition}) before detailing each component: the multi-granularity semantic representation (Section~\ref{sec:semantic_generation}), the adaptive partitioning and visual-semantic alignment (Section~\ref{sec:adaptive_fusion}), and the dynamic test-time feature refinement (Section~\ref{sec:dynamic_refinement}).

\subsection{Problem Definition}
\label{sec:problem_definition}
\noindent\textbf{Zero-Shot Learning (ZSL).}
The task of Zero-Shot Learning (ZSL)~\cite{pourpanah2022review} aims to enable models to recognize instances from classes that were not present during training. Formally, we are given a labeled training dataset $\mathcal{D}_s = \{(\mathbf{x}_i, y_i)\}_{i=1}^{N_s}$ from a set of \textit{seen} classes $\mathcal{Y}_s$, where $\mathbf{x}_i \in \mathcal{X}$ represents a visual input (in our case, a skeleton sequence), and $y_i \in \mathcal{Y}_s$ is its corresponding class label. The core challenge lies in enabling the model to correctly classify samples from a disjoint set of \textit{unseen} classes $\mathcal{Y}_u$ during inference, where $\mathcal{Y}_s \cap \mathcal{Y}_u = \emptyset$.

This cross-class generalization is made possible through a shared semantic space that bridges the gap between visual features and class concepts. Typically, this takes the form of pre-defined vector embeddings $\mathbf{F} = \{F_y \in \mathbb{R}^d : y \in \mathcal{Y}_s \cup \mathcal{Y}_u\}$, which are available for all classes (both seen and unseen). These embeddings encode semantic properties that are expected to correlate with visual characteristics. In the standard ZSL setting, the model is evaluated on a test set $\mathcal{D}_u = \{(\mathbf{x}_j, y_j)\}_{j=1}^{N_u}$ where $y_j \in \mathcal{Y}_u$, and the prediction space is restricted to unseen labels, i.e., $f: \mathcal{X} \rightarrow \mathcal{Y}_u$.

\noindent\textbf{Generalized Zero-Shot Learning (GZSL).}
While standard ZSL provides a useful theoretical framework, real-world applications often demand a more challenging setting known as Generalized ZSL (GZSL). In this setting, the test set contains samples from both seen and unseen classes, and the prediction space is expanded to the union of both label sets, $f: \mathcal{X} \rightarrow \mathcal{Y}_s \cup \mathcal{Y}_u$. This introduces a significant challenge: the model must not only correctly classify novel unseen actions but also maintain its ability to recognize seen actions it was trained on. Many ZSL approaches suffer from a strong bias towards seen classes in this setting, as the model has learned direct mappings for these classes during training, leading to a tendency to misclassify unseen samples as belonging to familiar seen categories.

\noindent\textbf{Test-Time Adaptation (TTA).}
Test-Time Adaptation (TTA)~\cite{wang2020tent,zhang2022memotesttimerobustness} represents a paradigm shift in how models handle distribution shifts between training and testing data. Rather than relying solely on the generalization capability learned during training, TTA enables models to dynamically adapt using the test data itself, typically in an online and unsupervised manner. For an incoming mini-batch of test data $\mathcal{B}_t = \{\mathbf{x}_i\}_{i=1}^B$ at time step $t$, the model parameters $\boldsymbol{\theta}$ are updated based on an adaptation objective $\mathcal{L}_{\text{adapt}}$:
\begin{equation}
  \boldsymbol{\theta}_t = \boldsymbol{\theta}_{t-1} - \alpha \nabla_{\boldsymbol{\theta}} \mathcal{L}_{\text{adapt}}(\mathcal{B}_t; \boldsymbol{\theta}_{t-1})
  \label{eq:tta_update}
\end{equation}
where $\alpha$ is the learning rate for adaptation. In the context of ZSL, TTA presents a particularly promising direction for bridging the gap between the training distribution $P_{\text{train}}(\mathbf{x} \mid y \in \mathcal{Y}_s)$ and the distinct test-time distribution $P_{\text{test}}(\mathbf{x} \mid y' \in \mathcal{Y}_u)$. This is especially crucial given that unseen classes may exhibit visual characteristics that differ systematically from those observed during training.

\subsection{Multi-Granularity Semantic Representation Generation Module}
\label{sec:semantic_generation}
Human action understanding inherently operates at multiple levels of granularity. When observing an action, we simultaneously process global patterns (e.g., "a person throwing something") and local details (e.g., "the arm extends upward" or "the legs provide stability"). Inspired by this multi-scale perception, we propose to decompose actions into semantically meaningful components that can be independently learned and transferred across different action classes.

\subsubsection{Fine-Grained Action Decomposition}
We regard an action as a specific combination of local body movements that can be spatially or temporally decomposed. Beyond label-level semantics, these local movements represent individual visual concepts that are transferable across different classes. For instance, the arm-raising motion in "shooting a basketball" shares visual similarities with the arm-raising in "hitting with an object," despite the different overall action contexts.

To systematically extract such underlying semantics, we leverage the powerful language understanding capabilities of GPT-3 to generate detailed textual descriptions for movements at different scales. We design three types of decomposition: (1) \textbf{Spatial Decomposition}, where we generate descriptions for movements performed by $P=4$ distinct body parts: `head', `hands', `torso', and `legs', as exemplified in Table~\ref{table:gpt3}; (2) \textbf{Temporal Decomposition}, where we divide the action into $Z=3$ contiguous temporal intervals: `start', `middle', and `end', as shown in Table~\ref{table:gpt3-2}; and (3) \textbf{Global Description}, where we generate holistic descriptions that augment the original action labels with more detailed semantic information about the overall motion pattern.

\begin{table*}[t]
\centering
\resizebox{.9\textwidth}{!}{%
\begin{tabular}{|c|cccc|}
\hline
\multirow{2}{*}{\textbf{Action}} & \multicolumn{4}{c|}{\textbf{Question: Describe in very short how each body part moves for \textless{}Action\textgreater{}.}} \\ \cline{2-5} 
& \multicolumn{1}{c|}{\textbf{Head}} & \multicolumn{1}{c|}{\textbf{Hands}} & \multicolumn{1}{c|}{\textbf{Torso}} & \textbf{Legs} \\ \hline
\begin{tabular}[c]{@{}c@{}}Hit another person \\ with something\end{tabular} &
\multicolumn{1}{c|}{\begin{tabular}[c]{@{}c@{}}Turn towards the \\ other person.\end{tabular}} &
\multicolumn{1}{c|}{\begin{tabular}[c]{@{}c@{}}Grip the object tightly \\ and thrust it forward.\end{tabular}} &
\multicolumn{1}{c|}{\begin{tabular}[c]{@{}c@{}}Twist and turn to generate \\ momentum for the strike.\end{tabular}} &
\begin{tabular}[c]{@{}c@{}}Stomp the ground to provide \\ additional force for the strike.\end{tabular} \\ \hline
\begin{tabular}[c]{@{}c@{}}Shoot at the \\ basket\end{tabular} &
\multicolumn{1}{c|}{\begin{tabular}[c]{@{}c@{}}Turn and look up \\ towards the basket.\end{tabular}} &
\multicolumn{1}{c|}{\begin{tabular}[c]{@{}c@{}}Grip the ball\\ and release it.\end{tabular}} &
\multicolumn{1}{c|}{\begin{tabular}[c]{@{}c@{}}Twist and extend to generate \\power for the shot.\end{tabular}} &
\begin{tabular}[c]{@{}c@{}}Bend slightly and \\ propel slightly upward.\end{tabular} \\ \hline
\end{tabular}%
}
\caption{Example body-part-based descriptions generated by GPT-3. The refined explanations reveal shared motion patterns between semantically related actions: both `hit another person with something' and `shoot at the basket' involve similar head turning and hand gripping movements, highlighting transferable local motion concepts.}
\label{table:gpt3}
\end{table*}

\begin{table*}[t]
\centering
\resizebox{.85\textwidth}{!}{%
\begin{tabular}{|c|ccc|c|}
\hline
\multirow{2}{*}{\textbf{Action}} &
\multicolumn{3}{c|}{\textbf{\begin{tabular}[c]{@{}c@{}}Question: Separate the motion of \textless{}Action\textgreater{} into three phases.\end{tabular}}} &
\multirow{2}{*}{\textbf{\begin{tabular}[c]{@{}c@{}}Question: Describe in very short \\ how a person does \textless{}Action\textgreater{}.\end{tabular}}} \\ \cline{2-4}
&
\multicolumn{1}{c|}{\textbf{Start}} &
\multicolumn{1}{c|}{\textbf{Middle}} &
\textbf{End} &
\\ \hline
\begin{tabular}[c]{@{}c@{}}Hit another person \\ with something\end{tabular} &
\multicolumn{1}{c|}{Raise arm.} &
\multicolumn{1}{c|}{Swing arm.} &
Strike other person. &
\begin{tabular}[c]{@{}c@{}}Swing their arm and strike the\\ other person with the object.\end{tabular} \\ \hline
\begin{tabular}[c]{@{}c@{}}Shoot at the \\ basket\end{tabular} &
\multicolumn{1}{c|}{Raise arm.} &
\multicolumn{1}{c|}{Throw ball.} &
Aim at basket. &
\begin{tabular}[c]{@{}c@{}}Raise their arm and throw the ball \\ towards the basket.\end{tabular} \\ \hline
\end{tabular}%
}
\caption{Example temporal-interval-based and global descriptions generated by GPT-3. The temporal decomposition reveals that both actions share a common starting phase (raising arm), demonstrating how temporal segments can capture transferable motion primitives across different action classes.}
\label{table:gpt3-2}
\end{table*}
\begin{figure}[t]
\centering
\includegraphics[width=.45\linewidth]{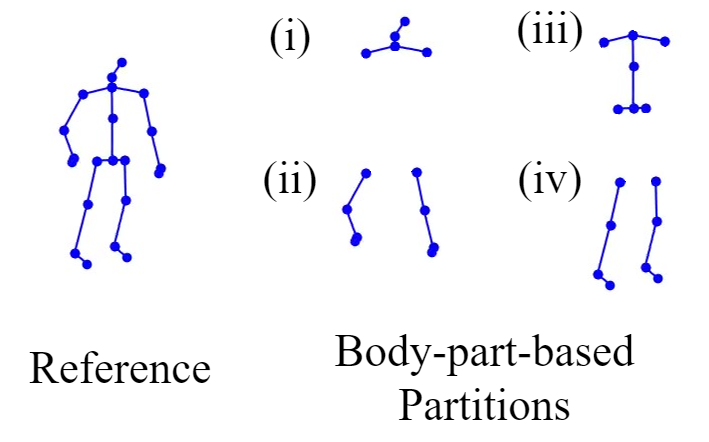}
\caption{Static spatial partitioning scheme for decomposing 25 body joints into four semantic body parts: (i) Head: joints 0, 1, 2, 3 (ii) Hands: joints 4-11, 21-24 (iii) Torso: joints 12, 16, 20 (iv) Legs: joints 13-15, 17-19.}
\label{fig:pt}
\vspace{-0.2cm}
\end{figure}

\subsubsection{Prompt Engineering and Embedding Generation}
To ensure consistent and informative responses from GPT-3, we employ a structured prompt template. For local part descriptions, we format our queries as: \texttt{Using the following format, <QUESTION>: <LOCAL PART 1> would: ...; <LOCAL PART 2> would: ...; \ldots; <LOCAL PART H> would: ...} where $H \in \{P, Z\}$ represents either the number of body parts or temporal segments. This structured format ensures that GPT-3 provides complete descriptions for all requested components in a consistent manner. The specific prompt content and outputs can be referenced in Tables~\ref{table:gpt3} and \ref{table:gpt3-2}.

After acquiring the targeted descriptions $\mathcal{D}_y = \{d_y^{(i)}\}_{i=0}^{P+Z}$ for each class $y$ (where $d_y^{(0)}$ represents the global description and $d_y^{(1)}$ through $d_y^{(P+Z)}$ represent local descriptions), we convert them into standard visual-language prompts. Specifically, each description is wrapped in a template such as "a video of [DESCRIPTION]" to maintain consistency with the pre-training objective of the visual-language model. These prompts are then processed through a pre-trained CLIP text encoder, $f_{\text{text}}$, to obtain their corresponding embeddings:
\begin{equation}
F_y^{(i)} = f_{\text{text}}(\mathrm{prompt}(d_y^{(i)})) \in \mathbb{R}^d
\end{equation}
where $\mathrm{prompt}(\cdot)$ denotes the prompting function and $d$ is the text embedding dimension. The embeddings are concatenated to form the static multi-granularity semantic anchor matrix for all $C$ classes:
\begin{equation}
\mathbf{F} = \{F_y : y \in \mathcal{Y}_s \cup \mathcal{Y}_u\} \in \mathbb{R}^{C \times (P+Z+1) \times d}
\end{equation}
This matrix $\mathbf{F}$ serves as the foundational semantic representation for our framework, encoding both global action concepts and local motion patterns that can be shared across different action classes.
\begin{figure}[t]
\centering
\includegraphics[width=0.95\linewidth]{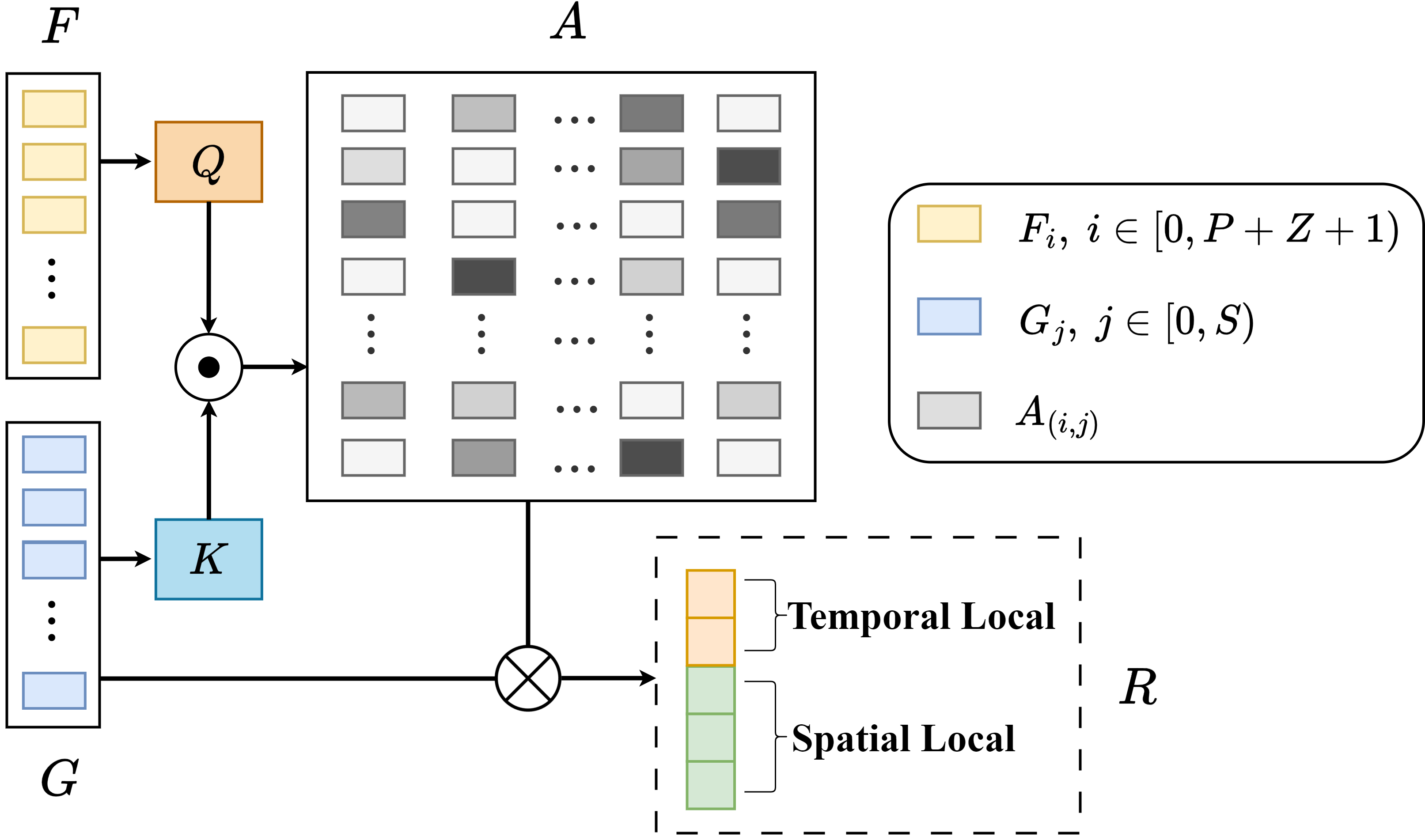}
\caption{Adaptive partitioning module. Each textual description acts as a query to attend over all spatio-temporal visual features, learning to focus on the most relevant nodes for that semantic concept.}
\label{fig:adaptive}
\vspace{-0.2cm}
\end{figure}

\subsection{Adaptive Partitioning and Visual-Semantic Alignment}
\label{sec:adaptive_fusion}
This section details our mechanism for adaptively partitioning and aligning multi-granularity semantic representations with visual features from skeleton sequences. This approach learns a robust alignment during training, which serves as the foundation for dynamic refinement at test time.

\subsubsection{Visual Feature Extraction}
Given a raw skeleton sequence $\mathbf{x} \in \mathbb{R}^{L\times J\times M\times 3}$, where $L, J, M$ are the sequence length, number of joints, and maximum number of persons, respectively, we first apply standard pre-processing (padding and normalization) following~\cite{shift-gcn}. We then employ a pre-trained Shift-GCN~\cite{shift-gcn} as our backbone encoder, denoted $g$, to extract rich spatio-temporal features:
\begin{equation}
\mathbf{G} = g(\mathbf{x}) \in \mathbb{R}^{S \times n}
\end{equation}
where $S = L' \times J$ is the total number of spatio-temporal nodes (with $L'$ being the temporal dimension after pooling), and $n$ is the visual feature dimension. Each row in $\mathbf{G}$ is a feature vector for a specific joint at a specific time, capturing both spatial configuration and temporal dynamics.

\subsubsection{Adaptive Partitioning via Cross-Modal Attention}
To align the visual feature map $\mathbf{G}$ with our multi-granularity text features $\mathbf{F}$, we require a mechanism to group visual features based on their semantic relevance to each description. A naive approach is static partitioning: manually grouping joints into pre-defined body parts (Fig.~\ref{fig:pt}) and dividing the sequence into fixed temporal segments. However, this method is fundamentally flawed due to its \textbf{rigidity}, as pre-defined partitions cannot adapt to action-specific motion variations. It also leads to a \textbf{loss of context} (e.g., leg motion is best understood with concurrent arm swing) and creates \textbf{boundary issues}, where hard boundaries can unnaturally sever semantically coherent motions.

To overcome these limitations, we propose an adaptive partitioning module based on cross-modal attention, depicted in Figure~\ref{fig:adaptive}. This approach learns to dynamically aggregate description-relevant features from the entire spatio-temporal map $\mathbf{G}$, allowing each semantic concept to attend to its most relevant visual evidence, irrespective of pre-defined boundaries. Importantly, this attention-based mechanism provides robustness to partial occlusions and viewpoint variations. When certain joints are occluded or exhibit weak motion, the attention weights can adaptively down-weight their contribution, while emphasizing more reliable features. This soft weighting scheme enables flexible handling of incomplete visual observations, as each textual description can focus on the most informative spatio-temporal regions for the current action.

Specifically, we treat the text embeddings as queries and the visual features as keys and values. The queries are learned projections of the static semantic embeddings:
\begin{equation}
  \mathbf{Q} = \mathbf{F}\mathbf{W}_Q \in \mathbb{R}^{(P+Z+1) \times h}
\end{equation}
where $\mathbf{W}_Q \in \mathbb{R}^{d \times h}$ is a learnable projection matrix. Similarly, the visual features are projected to form keys:
\begin{equation}
  \mathbf{K} = \mathbf{G}\mathbf{W}_K \in \mathbb{R}^{S \times h}
\end{equation}
where $\mathbf{W}_K \in \mathbb{R}^{n \times h}$ is another learnable projection. The attention matrix $\mathbf{A} \in \mathbb{R}^{(P+Z+1) \times S}$ is computed via scaled dot-product attention:
\begin{equation}
  \mathbf{A} = \mathrm{softmax}\left(\frac{\mathbf{Q}\mathbf{K}^{\top}}{\sqrt{h}}\right)
  \label{eq:attention}
\end{equation}
Intuitively, each row $\mathbf{A}_i$ of the attention matrix represents a soft selection of all $S$ spatio-temporal nodes with respect to the $i$-th textual description. The final fused visual representation $\mathbf{R} \in \mathbb{R}^{(P+Z+1) \times n}$ is obtained by a weighted aggregation of the visual features:
\begin{equation}
  \mathbf{R} = \mathbf{A}\mathbf{G}
  \label{eq:fused_visual}
\end{equation}
This attention-based fusion is \textbf{flexible}, \textbf{context-preserving}, and \textbf{learnable}, allowing it to adapt to the specific characteristics of each action class.

\subsubsection{Visual-Semantic Alignment and Training Objective}
The training objective is to learn a strong alignment between the fused visual representations and the static multi-granularity semantic embeddings. We project each row vector $\mathbf{r}_i \in \mathbb{R}^n$ from the fused visual representation $\mathbf{R}$ (Eq.~\ref{eq:fused_visual}) into the textual embedding space using a trainable MLP, $f_{\text{skel}}$:
\begin{equation}
\mathbf{v}_i = f_{\text{skel}}(\mathbf{r}_i) \in \mathbb{R}^d, \quad i \in \{0, 1, \ldots, P+Z\}
\end{equation}
We employ a symmetric InfoNCE loss~\cite{oord2018representation} to maximize the similarity between matched visual-text pairs while pushing negative pairs apart. The contrastive loss for a projected visual feature $\mathbf{v}_i$ and its ground-truth text feature $\mathbf{f}_i$ is:
\begin{equation}
\begin{aligned}
\mathcal{L}(\mathbf{v}_i, \mathbf{f}_i) = &-\frac{1}{2}\log\frac{\exp(\mathrm{sim}(\mathbf{v}_i, \mathbf{f}_i)/\tau)}{\sum_{o \in \mathcal{Y}_s}\exp(\mathrm{sim}(\mathbf{v}_i, \mathbf{f}_i^o)/\tau)} \\
&-\frac{1}{2}\log\frac{\exp(\mathrm{sim}(\mathbf{v}_i, \mathbf{f}_i)/\tau)}{\sum_{w \in \mathcal{B}}\exp(\mathrm{sim}(\mathbf{v}_i^w, \mathbf{f}_i)/\tau)}
\end{aligned}
 \label{eq:contrastive_loss}
\end{equation}
where $\mathrm{sim}(\cdot,\cdot)$ is cosine similarity, $\mathbf{f}_i^o$ are text embeddings from other (negative) seen classes, $\mathbf{v}_i^w$ are visual embeddings from other (negative) samples in the batch $\mathcal{B}$, and $\tau$ is a temperature parameter. The symmetric formulation ensures a robust bidirectional alignment.

The overall training loss aggregates the losses across all granularities using learnable importance weights $\alpha_i$:
\begin{equation}
\mathcal{L}_{\text{train}} (\mathbf{x}, y) = \sum_{i=0}^{P+Z}\alpha_i \mathcal{L}(\mathbf{v}_i, \mathbf{f}_i)
\label{eq:total_train_loss}
\end{equation}
These weights allow the model to adaptively balance the contribution of each semantic granularity, which is crucial as different actions rely on different scales of motion. The parameters optimized at this stage are $\boldsymbol{\theta}_{\text{train}} = \{g, \mathbf{W}_Q, \mathbf{W}_K, f_{\text{skel}}, \{\alpha_i\}\}$.

\subsection{Dynamic Test-Time Feature Refinement}
\label{sec:dynamic_refinement}
While the alignment module establishes a robust visual-semantic mapping, the static nature of text embeddings limits performance against test-time distribution shifts, especially for unseen classes. To address this, we introduce our primary innovation: a dynamic query refinement mechanism that adapts semantic representations online. This module builds upon the cross-modal attention from Section~\ref{sec:adaptive_fusion}. By refining the semantic features, which serve as queries, we indirectly yet effectively influence the visual feature aggregation, thereby dynamically adapting the final visual representations to the test distribution.

\begin{algorithm}[t]
\caption{DynaPURLS Inference Procedure}
\label{alg:tta}
\small  %
\setlength{\lineskip}{0pt}  %
\KwIn{Test samples $\{\mathbf{x}_i\}_{i=1}^{N_{\text{test}}}$, confidence threshold $\tau_{\text{conf}}$, bank capacity $K$, refinement rate $\beta$}
\KwOut{Predictions $\{\hat{y}_i\}_{i=1}^{N_{\text{test}}}$}
Initialize $\mathcal{M}_c = \emptyset$ for each class $c$, $\mathbf{S} = \mathbf{1}$, $\Delta\mathbf{F} = \mathbf{0}$\;
\For{each test sample $\mathbf{x}_i$}{
  $\mathbf{G}_i = g(\mathbf{x}_i)$, $\mathbf{F}' = \mathcal{N}(\mathbf{S} \odot \mathbf{F} + \Delta\mathbf{F})$, $\mathbf{Q}' = \mathbf{F}'\mathbf{W}_Q$\;
  $\mathbf{R}_i = \mathrm{Attention}(\mathbf{Q}', \mathbf{G}_i)$, $\mathbf{v}_i = f_{\text{skel}}(\mathbf{r}_{i,\text{global}})$\;
  Compute $\mathbf{p}_i$ using $\mathbf{v}_i$ and $\mathbf{F}'$, $\hat{y}_i = \argmax_c p_{i,c}$, $\mathrm{conf}_i = p_{i,\hat{y}_i}$\;
  
  \If{$\mathrm{conf}_i > \tau_{\text{conf}}$}{
    Update $\mathcal{M}_{\hat{y}_i}$ with $(\mathbf{x}_i, \hat{y}_i, \mathrm{conf}_i)$\;
    \If{$|\mathcal{M}_{\hat{y}_i}| > K$}{Remove sample with lowest confidence from $\mathcal{M}_{\hat{y}_i}$\;}
  }
  
  \If{$|\mathcal{M}| \geq B_{\text{min}}$}{
    Sample class-balanced batch $\mathcal{B} \subset \mathcal{M}$\;
    $\mathcal{L}_{\text{adapt}} = -\frac{1}{|\mathcal{B}|} \sum_{(\mathbf{x}_j, \hat{y}_j) \in \mathcal{B}} \log p'_{j,\hat{y}_j}$\;
    $\mathbf{S} \leftarrow \mathbf{S} - \beta \nabla_{\mathbf{S}} \mathcal{L}_{\text{adapt}}$, $\Delta\mathbf{F} \leftarrow \Delta\mathbf{F} - \beta \nabla_{\Delta\mathbf{F}} \mathcal{L}_{\text{adapt}}$\;
  }
}
\Return{$\{\hat{y}_i\}_{i=1}^{N_{\text{test}}}$}
\end{algorithm}

\subsubsection{Dynamic Semantic Refinement}
Instead of using static queries from $\mathbf{F}$, we introduce lightweight, learnable parameters to adjust the semantic embeddings based on test data characteristics. For each test batch, the refined text features $\mathbf{F}'$ are computed via an affine transformation:
\begin{equation}
  \mathbf{F}' = \mathcal{N}\left(\mathbf{S} \odot \mathbf{F} + \Delta\mathbf{F}\right)
  \label{eq:refinement_detailed}
\end{equation}
where $\mathbf{S} \in \mathbb{R}^{C \times (P+Z+1) \times d}$ is a learnable scaling tensor, $\Delta\mathbf{F} \in \mathbb{R}^{C \times (P+Z+1) \times d}$ is a learnable bias tensor, $\odot$ denotes element-wise multiplication, and $\mathcal{N}(\cdot)$ is per-vector $L_2$ normalization. This provides an effective yet efficient mechanism for adapting embeddings to the test distribution.

The test-time fusion process then uses dynamic queries constructed from these refined features:
\begin{equation}
  \mathbf{Q}' = \mathbf{F}'\mathbf{W}_Q
\end{equation}
The subsequent attention and feature aggregation proceed as before, but now with adaptive queries. This dynamic adjustment of semantic queries indirectly refines the fused skeleton features by altering the attention weights. This refinement is particularly valuable when test-time visual observations differ from training expectations due to occlusions, viewpoint variations, or domain shifts. By adapting the semantic anchors online, the model can help reduce misalignments between idealized text descriptions and actual visual evidence, thereby improving cross-modal alignment robustness under imperfect conditions. Critically, during this stage, the main network parameters $\boldsymbol{\theta}_{\text{train}}$ are frozen. Only the lightweight refinement parameters $\boldsymbol{\theta}_{\text{adapt}} = \{\mathbf{S}, \Delta\mathbf{F}\}$ are updated, ensuring efficient refinement without catastrophic forgetting.

\subsubsection{Confidence-Guided Online Optimization}
The efficacy of TTA hinges on the quality of its self-supervisory signal. Naive entropy minimization often leads to confirmation bias, where the model reinforces its own incorrect predictions. To avert this, we propose a more robust strategy based on selective optimization.

\noindent\textbf{Class-Balanced Memory Bank.} Our key insight is that samples with high prediction confidence provide more reliable supervisory signals. We introduce a \textbf{Class-Balanced Memory Bank} $\mathcal{M}$ to store a balanced set of high-confidence, pseudo-labeled test samples. As detailed in Algorithm~\ref{alg:tta}, for each incoming sample $\mathbf{x}_i$, we compute its prediction probability distribution using the current refined semantic embeddings $\mathbf{F}'$:
\begin{equation}
 p_{i,c} = \frac{\exp\left( \mathrm{sim}(\mathbf{v}_i, \mathbf{f}'_c) / \tau \right)}{\sum_{j=1}^C \exp\left( \mathrm{sim}(\mathbf{v}_i, \mathbf{f}'_j) / \tau \right)}
\end{equation}
where $\mathbf{v}_i$ is the global visual feature and $\mathbf{f}'_c$ is the refined global text feature for class $c$.

The memory bank maintains at most $K$ samples per class, preventing bias towards frequent classes. When a class's bank is full, new high-confidence samples replace those with the lowest confidence, progressively improving the quality of stored exemplars.

\noindent\textbf{Refinement Objective.} Using the curated samples in $\mathcal{M} = \bigcup_c \mathcal{M}_c$, we define the refinement loss as the cross-entropy over the pseudo-labels:
\begin{equation}
 \mathcal{L}_{\text{adapt}} = -\frac{1}{|\mathcal{B}|} \sum_{(\mathbf{x}_j, \hat{y}_j) \in \mathcal{B}} \log p'_{j,\hat{y}_j}
 \label{eq:adaptation_loss_detailed}
\end{equation}
where $p'_{j,\hat{y}_j}$ is the probability for sample $\mathbf{x}_j$ from a balanced batch $\mathcal{B} \subset \mathcal{M}$, recomputed with the current adaptive parameters. These parameters are then updated via gradient descent:
\begin{equation}
 \boldsymbol{\theta}_{\text{adapt}} \leftarrow \boldsymbol{\theta}_{\text{adapt}} - \beta \nabla_{\boldsymbol{\theta}_{\text{adapt}}} \mathcal{L}_{\text{adapt}}
 \label{eq:update_params_detailed}
\end{equation}
where $\beta$ is the refinement learning rate. This selective optimization process refines the semantic embeddings to better align with the true test distribution, stabilizing adaptation and improving generalization to unseen classes.

\subsection{GZSL Protocol Extension}
\label{sec:gzsl_extension}
In GZSL, the model must differentiate between seen and unseen classes. Inspired by OOD detection methods~\cite{yuan2024discriminability,Gao_2024_CVPR,chan2021entropy}, we implement an entropy-based gating mechanism. We first calculate the predictive entropy for each test sample $\mathbf{x}_i$ with respect to the \textit{entire} set of class semantics $\mathcal{Y}_s \cup \mathcal{Y}_u$:
\begin{equation}
  H(\mathbf{p}_i) = - \sum_{c=1}^{C} p_{i,c} \log p_{i,c}.
\end{equation}
Using an entropy threshold $\delta$ tuned on a validation set, we triage samples: those with entropy below $\delta$ (low uncertainty) are classified as likely `seen', while those above $\delta$ (high uncertainty) are deemed `unseen'. Following this separation, we classify each subset independently using only the corresponding `seen' or `unseen' class prototypes. This prevents the model from defaulting to high-confidence seen class predictions for unseen samples. The class-balanced memory bank is especially critical here, as it maintains distinct, high-quality representations for both seen and unseen classes, preventing feature drift and ensuring the refinement process remains robust.

\noindent\textbf{Summary.} Our DynaPURLS framework addresses the semantic drift challenge in zero-shot skeleton-based action recognition through: (1) multi-granularity semantic representations that capture transferable motion concepts, (2) adaptive partitioning and visual-semantic alignment that flexibly aligns features across modalities and establishes robust multi-scale correspondences, and (3) dynamic test-time feature refinement that adapts semantic embeddings and indirectly refines visual features, with entropy-based domain separation for robust GZSL performance. Together, these components enable effective knowledge transfer to unseen action classes while maintaining strong performance on seen classes.

\section{Experiments}
\label{sec:experiment}

\subsection{Experimental Setup}
\label{subsec:setup}

\subsubsection{Datasets}\label{subsubsec:datasets}
We evaluate our approach on three widely-used, large-scale benchmarks for skeleton-based action recognition to ensure a comprehensive assessment of its capabilities.

\textbf{NTU RGB+D 60}~\cite{shahroudy2016ntu} is a foundational dataset in the field, containing 56,880 skeleton sequences that cover 60 distinct action classes. These actions were performed by 40 different subjects and captured from 80 viewpoints, providing significant diversity. Each sequence offers 3D coordinates for 25 primary body joints and can accommodate up to two performers, with coordinate values being padded with zeros for single-person actions.

\textbf{NTU RGB+D 120}~\cite{liu2019ntu} serves as a large-scale extension of its predecessor, significantly increasing the complexity and diversity with 114,480 samples across 120 action classes. The data was collected from a larger pool of 106 subjects and across 155 distinct viewpoints, presenting a more challenging scenario for generalization. It maintains the same 25-joint skeleton representation, making it directly comparable with NTU RGB+D 60.

\textbf{PKU-MMD}~\cite{liu2017pku} is another critical multi-modal benchmark, comprising approximately 20,000 action instances that span 51 categories. A key feature of this dataset is its division into two phases of increasing difficulty (Phase I and II), allowing for evaluation under varying levels of challenge. Although it follows the same 25-joint skeleton format as the NTU RGB+D datasets, it introduces different action types and recording conditions, making it an excellent testbed for evaluating model robustness and cross-dataset generalization.

\subsubsection{Evaluation Protocols}\label{subsubsec:evaluation_protocols}
We conduct experiments under two standard protocols. In \textbf{Zero-Shot Learning (ZSL)}, the model is trained exclusively on seen classes and evaluated only on unseen classes, where we report Top-1 accuracy on the unseen test set. In \textbf{Generalized Zero-Shot Learning (GZSL)}, the model must distinguish between both seen and unseen classes during testing, presenting a more realistic but challenging scenario. For GZSL, we report accuracy for seen classes (S), unseen classes (U), and their harmonic mean (H = $\frac{2 \times S \times U}{S + U}$), where H provides a balanced measure of performance across both domains.

Following established practices \cite{gupta2021syntacticallyguidedgenerativeembeddings}, we employ multiple class split configurations to comprehensively assess generalization capability. For NTU RGB+D 60, we use 55/5 and 48/12 splits (seen/unseen classes), while NTU RGB+D 120 uses 110/10 and 96/24 splits. Additionally, recent works \cite{Zhou_2023, li2024sadvaeimprovingzeroshotskeletonbased} have introduced random split settings to provide more robust evaluation. For fair comparison, we adopt the same three public random splits from prior work for each dataset: three random 55/5 splits for NTU 60, three random 110/10 splits for NTU 120, and three random 46/5 splits for PKU-MMD (using all 51 categories from both Phase I and II \cite{Zhou_2023, chen2024finegrainedinformationguideddualprompts}). We report averaged results across all random trials to ensure consistency with previous evaluations.

\subsubsection{Baseline Methods}\label{subsubsec:baseline_methods}
We compare against comprehensive state-of-the-art skeleton-based zero-shot learning methods spanning different paradigms. The embedding-based methods include DeViSE \cite{devise}, which learns a linear transformation from visual to semantic space; ReViSE \cite{hubert2017learning}, which employs bidirectional projections between modalities; and JPoSE \cite{wray2019fine}, which leverages part-of-speech embeddings for fine-grained action understanding. Among generative approaches, CADA-VAE \cite{schonfeld2019generalized} uses variational autoencoders to synthesize visual features from semantic descriptions, while SynSE \cite{gupta2021syntacticallyguidedgenerativeembeddings} exploits syntactic structures in action labels to guide feature generation.

Recent skeleton-specific methods have introduced specialized architectures. These include SMIE \cite{Zhou_2023}, which maximizes mutual information between visual and textual distributions; SA-DVAE \cite{li2024sadvaeimprovingzeroshotskeletonbased}, which employs disentangled autoencoders to separate action-specific features; STAR \cite{chen2024finegrainedinformationguideddualprompts}, which utilizes dual prompts guided by fine-grained information; Neuron \cite{chen2024neuronlearningcontextawareevolving}, which introduces a framework for learning context-aware evolving representations; and SCoPLe \cite{Zhu_2025_CVPR}, which pioneers semantic-guided cross-modal prompt learning. We also extend our PURLS \cite{zhu2024partawareunifiedrepresentationlanguage} method to GZSL and other experimental settings to facilitate comprehensive evaluation and demonstrate the consistent improvements achieved by DynaPURLS across different scenarios.

\begin{table*}[t]  
    \caption{Comparison with state-of-the-art methods on NTU RGB+D 60, NTU RGB+D 120, and Kinetics-skeleton 200 under fixed splits. \textbf{ZSL Acc} denotes the Top-1 accuracy in the ZSL setting (highlighted in gray). \textbf{S}, \textbf{U}, and \textbf{H} denote the seen accuracy, unseen accuracy, and harmonic mean in the GZSL setting, respectively. \textbf{H} is highlighted in gray as the key metric. \textbf{Bold} indicates the best performance, \underline{underline} indicates the second best.}  
    \label{table:comparison}  
    \centering  
    \footnotesize  
    \setlength{\tabcolsep}{3.5pt}  
    \resizebox{\textwidth}{!}{  
    \begin{tabular}{l >{\columncolor{lightgray}}c cc>{\columncolor{lightgray}}c >{\columncolor{lightgray}}c cc>{\columncolor{lightgray}}c >{\columncolor{lightgray}}c cc>{\columncolor{lightgray}}c >{\columncolor{lightgray}}c cc>{\columncolor{lightgray}}c >{\columncolor{lightgray}}c cc>{\columncolor{lightgray}}c >{\columncolor{lightgray}}c cc>{\columncolor{lightgray}}c}  
        \toprule  
        \multirow{3}{*}{Method} & \multicolumn{8}{c}{NTU RGB+D 60} & \multicolumn{8}{c}{NTU RGB+D 120} & \multicolumn{8}{c}{Kinetics-skeleton 200} \\
        \cmidrule(lr){2-9} \cmidrule(lr){10-17} \cmidrule(lr){18-25}  
         & \multicolumn{4}{c}{55/5} & \multicolumn{4}{c}{48/12} & \multicolumn{4}{c}{110/10} & \multicolumn{4}{c}{96/24} & \multicolumn{4}{c}{180/20} & \multicolumn{4}{c}{160/40} \\
        \cmidrule(lr){2-5} \cmidrule(lr){6-9} \cmidrule(lr){10-13} \cmidrule(lr){14-17} \cmidrule(lr){18-21} \cmidrule(lr){22-25}  
         & ZSL  & S & U & H & ZSL  & S & U & H & ZSL  & S & U & H & ZSL & S & U & H & ZSL  & S & U & H & ZSL  & S & U & H \\
        \midrule  
        ReViSE \cite{hubert2017learning} & 53.91 & 74.22 & 34.73 & 47.32 & 17.49 & 62.36 & 20.77 & 31.16 & 55.04 & 48.69 & 44.84 & 46.68 & 32.38 & 49.66 & 25.06 & 33.31 & 24.95 & 24.52 & 18.96 & 21.28 & 13.28 & 24.61 & 10.09 & 14.33 \\
        JPoSE \cite{wray2019fine} & 64.82 & 64.44 & 50.29 & 56.49 & 28.75 & 60.49 & 20.62 & 30.75 & 51.93 & 47.66 & 46.40 & 47.05 & 32.44 & 38.62 & 22.79 & 28.67 & 28.40 & 25.58 & 22.44 & 23.89 & 16.75 & 25.47 & 13.23 & 17.38 \\
        CADA-VAE \cite{schonfeld2019generalized} & 76.84 & 69.38 & 61.79 & 65.37 & 28.96 & 51.32 & 27.03 & 35.41 & 59.53 & 47.16 & 49.78 & 48.44 & 35.77 & 41.11 & 34.14 & 37.31 & 27.15 & 26.14 & 20.09 & 22.71 & 15.82 & 26.23 & 11.71 & 16.18 \\
        SynSE \cite{gupta2021syntacticallyguidedgenerativeembeddings} & 75.81 & 61.27 & 56.93 & 59.02 & 33.30 & 52.21 & 27.85 & 36.33 & 62.69 & 52.51 & 57.60 & 54.94 & 38.70 & 56.39 & 32.25 & 41.04 & - & - & - & - & - & - & - & - \\
        SMIE \cite{Zhou_2023} & 77.98 & - & - & - & 40.18 & - & - & - & 65.74 & - & - & - & 45.30 & - & - & - & - & - & - & - & - & - & - & - \\
        STAR \cite{chen2024finegrainedinformationguideddualprompts} & 81.40 & 69.00 & 69.90 & 69.40 & 45.10 & 62.70 & 37.00 & 46.60 & 63.30 & 59.90 & 52.70 & 56.10 & 44.30 & 51.20 & 36.90 & 42.90 & - & - & - & - & - & - & - & - \\
        SA-DVAE \cite{li2024sadvaeimprovingzeroshotskeletonbased} & 82.37 & 62.28 & 70.80 & 66.27 & 41.38 & 50.20 & 36.94 & 42.56 & 68.77 & 61.10 & 59.75 & 60.42 & 46.12 & 58.82 & 35.79 & 44.50 & - & - & - & - & - & - & - & - \\
        Neuron \cite{chen2024neuronlearningcontextawareevolving} & \underline{86.90} & 69.10 & 73.80 & \textbf{71.40} & \underline{62.70} & 61.60 & 56.80 & \underline{59.10} & 71.50 & 67.60 & 59.50 & 63.30 & \underline{57.10} & 67.50 & 44.40 & 53.60 & - & - & - & - & - & - & - & - \\
        SCoPLe \cite{Zhu_2025_CVPR} & 84.10 & 69.60 & 71.94 & \underline{70.75} & 52.96 & 54.49 & 61.83 & 57.93 & \underline{74.53} & 63.51 & 61.08 & 62.27 & 52.17 & 53.33 & 51.18 & 52.24 & - & - & - & - & - & - & - & - \\
        \midrule  
        PURLS \cite{zhu2024partawareunifiedrepresentationlanguage} & 79.22 & 71.70 & 60.35 & 65.53 & 40.99 & 81.60 & 36.92 & 50.84 & 71.95 & 76.25 & 67.89 & \underline{71.80} & 52.01 & 72.67 & 45.32 & \underline{55.81} & \underline{32.22} & 26.77 & 26.09 & \underline{26.43} & \underline{22.56} & 26.89 & 18.27 & \underline{21.84} \\
        \textbf{DynaPURLS (Ours)} & \textbf{88.52} & 72.67 & 67.43 & 69.95 & \textbf{71.80} & 82.48 & 61.81 & \textbf{70.66} & \textbf{89.06} & 80.02 & 83.00 & \textbf{81.49} & \textbf{69.11} & 74.41 & 56.70 & \textbf{64.36} & \textbf{40.52} & 27.49 & {43.68} & \textbf{33.76} & \textbf{35.08} & 27.58 & {36.96} & \textbf{31.67} \\
        \bottomrule  
    \end{tabular}  
    }  
\end{table*}

\begin{table*}[t]
    \caption{Comparison with state-of-the-art methods on NTU RGB+D 60, NTU RGB+D 120, and PKU-MMD under random split settings. \textbf{ZSL} denotes the Top-1 accuracy in the ZSL setting. \textbf{S}, \textbf{U}, and \textbf{H} denote the seen accuracy, unseen accuracy, and harmonic mean in the GZSL setting, respectively. \textbf{Bold} indicates the best performance, \underline{underline} indicates the second best.}
    \label{table:comparison2}
    \centering
    \setlength{\tabcolsep}{3pt}
    
    \footnotesize
    \begin{tabular}{l cccc cccc cccc}
        \toprule
        \multirow{2}{*}{Method} & \multicolumn{4}{c}{NTU RGB+D 60} & \multicolumn{4}{c}{NTU RGB+D 120} & \multicolumn{4}{c}{PKU-MMD} \\
        \cmidrule(lr){2-5} \cmidrule(lr){6-9} \cmidrule(lr){10-13}
        & ZSL & S & U & H & ZSL & S & U & H & ZSL & S & U & H \\
        \midrule
        ReViSE \cite{hubert2017learning} & 60.94 & 71.75 & 52.06 & 60.34 & 44.90 & 48.29 & 34.64 & 40.34 & 59.34 & 60.89 & 42.16 & 49.82 \\
        JPoSE \cite{wray2019fine} & 59.44 & 66.25 & 54.92 & 60.05 & 46.69 & 49.43 & 39.14 & 43.69 & 57.17 & 60.26 & 45.18 & 51.64 \\
        CADA-VAE \cite{schonfeld2019generalized} & 61.84 & 77.35 & 58.14 & 66.38 & 45.15 & 51.09 & 41.24 & {45.64} & 60.74 & 63.17 & 35.86 & 45.75 \\
        SynSE \cite{gupta2021syntacticallyguidedgenerativeembeddings} & 64.19 & 75.84 & 60.77 & 67.47 & 47.28 & 41.73 & 45.36 & 43.47 & 53.85 & 63.09 & 40.69 & 49.47 \\
        SMIE \cite{Zhou_2023} & 65.08 & - & - & - & 46.40 & - & - & - & 60.83 & - & - & - \\
        SA-DVAE \cite{li2024sadvaeimprovingzeroshotskeletonbased} & \underline{84.20} & 78.16 & 72.60 & 75.27 & 50.67 & 58.09 & 40.23 & 47.54 & 66.54 & 58.49 & 51.40 & 54.72 \\
        SCoPLe \cite{Zhu_2025_CVPR} & 83.72 & 75.32 & 80.17 & \underline{77.67} & \underline{53.34} & 70.47 & 44.29 & \underline{54.08} & \underline{71.41} & 62.17 & 49.69 & \underline{54.85} \\
        \midrule

        \textbf{DynaPURLS (Ours)} & \textbf{86.75} & 86.68 & 73.50 & \textbf{79.44} & \textbf{90.04} & 76.80 & 72.80 & \textbf{74.72} & \textbf{78.26} & 54.88 & 66.50 & \textbf{60.11} \\
        \bottomrule
    \end{tabular}
\end{table*}

\subsubsection{Implementation Details}\label{subsubsec:implementation_details}
Our framework is meticulously implemented, building upon our proposed multi-granularity architecture with a novel dynamic inference-time refinement mechanism. The visual backbone is a \textbf{Shift-GCN}~\cite{cheng2020skeleton}, which has proven effective for skeleton-based action recognition. It extracts rich 256-dimensional spatio-temporal features from input skeleton sequences through its efficient shift graph convolution and temporal convolution operations. Following the standard ZSL protocol to prevent any information leakage, we pre-train this visual encoder using only the seen class samples specific to each experimental split configuration~\cite{synse-zsl}. All training parameters are kept consistent with those used in~\cite{synse-zsl} and~\cite{Zhou_2023} to ensure fair comparison.

For the generation of semantic representations, we employ \textbf{GPT-3 (text-davinci-003)} language model. We use carefully engineered prompts to guide the model to produce detailed and structured action descriptions. These prompts systematically decompose each action into spatial components (covering 4 body parts: head, hands, torso, and legs) and temporal phases (across 3 segments: start, middle, and end). The resulting textual descriptions are then encoded into high-quality 512-dimensional feature vectors using the pre-trained \textbf{CLIP text encoder (ViT-B/32)}. The core of our adaptive alignment is the cross-modal attention mechanism, which employs learnable projection matrices $W_Q \in \mathbb{R}^{512\times 150}$ and $W_K \in \mathbb{R}^{256\times 150}$, with a shared hidden dimension of $h = 150$. The final visual representation is projected into the semantic space by a projection head, $f_{skel}$, which consists of a 2-layer MLP with 512 hidden units and ReLU activation functions.

Our training is configured for robust convergence. We use the Adam optimizer with a learning rate of $1 \times 10^{-4}$ and a batch size of 256. The model is trained for a maximum of 300 epochs, but we employ an early stopping strategy that halts training if the validation accuracy does not improve for 20 consecutive epochs. For data processing, sequences from the NTU datasets are processed with a length of $L = 300$ and $J = 25$ joints, accommodating up to $M = 2$ performers. For the PKU-MMD dataset, we follow the configuration from~\cite{st-gcn} with $J = 18$ joints. 

At inference time, our dynamic refinement mechanism optimizes the lightweight affine transformation parameters $\mathcal{S}$ and $\Delta\mathcal{F}$. This is done using an Adam optimizer with a higher initial learning rate of 0.01, which is then decayed using a cosine annealing schedule. The class-balanced memory bank, which is central to our method's stability, is configured to maintain up to $K = 16$ high-confidence samples per class, selected based on a confidence threshold of $\tau = 0.1$. All reported results use 10 refinement steps at inference, organized into 4 progressive stages (steps 1-3, 4-5, 6-8, and 9-10), allowing the model to gradually adapt semantic features to the incoming visual stream. The adaptive parameters ($\mathcal{S} \in \mathbb{R}^{C \times 8 \times 512}$ and $\Delta\mathcal{F} \in \mathbb{R}^{C \times 8 \times 512}$) comprise only 8,192 parameters per class, confirming the lightweight nature of our approach. 

Table~\ref{tab:refinement_overhead} details the memory and computational costs of our dynamic refinement module. The learnable parameters ($\mathcal{S}$ and $\Delta\mathcal{F}$, each $\mathbb{R}^{C \times 8 \times 512}$) require minimal memory from 0.16 MB to 0.75 MB. The memory bank stores up to K=16 high-confidence samples per class with refined semantic features ($\mathbb{R}^{8 \times 512}$), requiring 2.13-9.68 MB. The 10-step refinement process requires only 1.0-7.5 seconds for the entire test set, demonstrating practical efficiency for real-world deployment.

\begin{table}[t]
\centering
\footnotesize
\caption{Memory and computational overhead of dynamic refinement across different dataset splits. All measurements are for the ZSL setting with 10 refinement steps on a single NVIDIA A100 GPU.}
\label{tab:refinement_overhead}
\begin{tabular}{l c c c c c}
\toprule
\textbf{Split} & \textbf{Unseen} & \textbf{Per-Class} & \textbf{Total} & \textbf{Bank} & \textbf{Time} \\
 & \textbf{Classes} & \textbf{(KB)} & \textbf{(MB)} & \textbf{(MB)} & \textbf{(s)} \\
\midrule
NTU60 55/5 & 5 & 32.8 & 0.16 & 2.13 & 1.0 \\
NTU60 48/12 & 12 & 32.5 & 0.38 & 4.87 & 2.5 \\
NTU120 110/10 & 10 & 32.8 & 0.32 & 4.26 & 3.2 \\
NTU120 96/24 & 24 & 32.4 & 0.75 & 9.68 & 7.5 \\
\bottomrule
\end{tabular}
\end{table}

Test batches use a size of 1 to mimic real online scenarios for efficient processing. All experiments reported were conducted on a single NVIDIA A100 GPU using PyTorch 1.12.

\subsection{Comparison with State-of-the-art Methods}
\label{subsec:comparison}

DynaPURLS achieves competitive results across all benchmarks, as shown in Tables~\ref{table:comparison} and~\ref{table:comparison2}. Following recent methods (SA-DVAE, Neuron, SCoPLe), we focus on NTU RGB+D 60/120 as standard benchmarks, with additional Kinetics-skeleton 200 results for comparison with earlier works.

In the ZSL setting (Table~\ref{table:comparison}), our method demonstrates strong performance, particularly on large-scale datasets. On NTU60, DynaPURLS achieves 88.52\% and 71.80\% (55/5, 48/12 splits), surpassing Neuron by 1.62\% and 9.1\%. The performance advantage is also observed on the complex NTU120 dataset, where we achieve 89.06\% (110/10) and 69.11\% (96/24), outperforming SCoPLe and Neuron by margins of 14.53\% and 12.01\%, respectively. This improvement on NTU120 likely stems from the method's ability to handle greater complexity and higher inter-class similarity through dynamic refinement, which effectively resolves ambiguities that static representations might struggle to capture. On random splits (Table~\ref{table:comparison2}), we achieve 90.04\% ZSL accuracy on NTU120, further validating the robustness of our approach.

The advantages of DynaPURLS extend to the generalized Zero-Shot Learning (GZSL) setting, where it achieves higher harmonic means (H) in three of four fixed splits, with distinct gains of up to 11.56\%. Notably, on random splits, we attain a 74.72\% H-score on NTU120 and 60.11\% on PKU-MMD. In these GZSL scenarios, DynaPURLS appears to maintain a balanced recognition capability. By refining semantic features at test time, our method aims to alleviate the inherent domain shift in ZSL, ensuring that unseen class representations remain distinctive.

This performance across diverse settings can be attributed to our dual innovation: multi-granularity representations combined with dynamic refinement. While traditional methods rely on static semantic-visual alignments, our approach actively attempts to bridge the semantic-visual gap. The dynamic refinement mechanism adapts the semantic space to match incoming visual evidence, correcting local misalignments, such as specific body part motions, without disrupting global structure. Stabilized by a class-balanced memory bank, this adaptive process helps the framework generalize effectively even in challenging, large-scale scenarios like NTU120, suggesting a benefit in moving from static to adaptive zero-shot recognition.

\section{Ablation Studies \& Qualitative Analysis}\label{sec:ablation}

To thoroughly understand the effectiveness of our approach and the contribution of each component, we conduct extensive ablation studies and qualitative analysis. This comprehensive evaluation reveals the effectiveness of individual components in both PURLS and DynaPURLS, as well as the underlying principles behind their success.

\subsection{Component-Level Ablations}\label{subsec:component_ablations}

This section empirically validates the architectural design choices, assessing both framework universality and the efficacy of the multi-granularity alignment strategy.

\noindent\textbf{Framework Universality.}
We first examine whether our multi-granularity alignment strategy generalizes across different visual backbones and language models. In this experiment, we compare our full model against a \textbf{`Global`} baseline, which aligns globally-averaged visual features with a single holistic action description (generated by the specified language model), lacking the multi-granularity body-part and temporal decomposition of our approach. We vary the skeleton encoders (AA-GCN, CTR-GCN, DG-GCN, PoseC3D, Shift-GCN) and the action descriptor generators (GPT-3, GPT-3.5, GPT-4), with results presented in Table~\ref{table:universality}.

The results demonstrate the robust generality of our approach. Our method achieves consistent and substantial performance gains, up to 18\% in absolute accuracy over the global baseline on the standard 55/5 split, regardless of the architectural configuration. Crucially, the performance variance across different language models (2.3\% and 0.9\% between GPT-3 and GPT-4 on the two splits) is far smaller than the consistent improvements our framework achieves, demonstrating that the effectiveness of our approach is fundamentally driven by the adaptive alignment and dynamic refinement mechanisms rather than semantic descriptor quality. Notably, our framework successfully adapts to PoseC3D, which outputs 2D heatmap features rather than graph-structured data. While a static partitioning scheme would fail in this scenario, our adaptive attention mechanism naturally handles the different input structure by learning pixel-wise attention weights, validating the flexibility and universality of our design. We adopt GPT-3 across all experiments for consistency with prior work and to ensure fair comparison with baseline methods that predominantly use GPT-3.

\begin{table}[t]
\centering
\footnotesize
\caption{Ablation study on NTU RGB+D 60 (\%) examining framework universality by replacing the skeleton encoder backbone or action descriptor generator.}
\label{table:universality}
\begin{tabular}{@{}cccccc@{}}
\toprule
\multirow{2}{*}{\textbf{Encoder}} & \multirow{2}{*}{\textbf{Descriptor}} & \multirow{2}{*}{\textbf{Model}} & \multicolumn{2}{c}{\textbf{NTU RGB+D 60 }} \\
& & & 55/5 & 48/12 \\ \midrule
AA \cite{aa-gcn} & GPT3 & Global & 62.79 & 28.09 \\
AA \cite{aa-gcn} & GPT3 & Ours & 76.75 & 32.39 \\
CTR \cite{ctr-gcn} & GPT3 & Global & 65.16 & 34.56 \\
CTR \cite{ctr-gcn} & GPT3 & Ours & 79.97 & 39.42 \\
DG \cite{dg-gcn} & GPT3 & Global & 64.28 & 34.04 \\
DG \cite{dg-gcn} & GPT3 & Ours & 80.41 & 41.06 \\
PoseC3D \cite{c3d} & GPT3 & Global & 63.45 & 35.71 \\
PoseC3D \cite{c3d} & GPT3 & Ours & 81.14 & {41.60} \\
Shift & GPT3 & Global & 64.69 & 35.46 \\
Shift & GPT3 & Ours & 79.23 & 40.99 \\
Shift & GPT3.5 & Global & 66.49 & 38.01 \\
Shift & GPT3.5 & Ours & 79.17 & 40.98 \\
Shift & GPT4 & Global & 64.71 & 40.76 \\
Shift & GPT4 & Ours & \textbf{81.53} & \textbf{41.90} \\ \bottomrule
\end{tabular}
\end{table}

\noindent\textbf{Semantic Enrichment and Partitioning.} This study analyzes the impact of semantic representations and partitioning strategies on performance, both before and after applying our inference-time dynamic refinement. We evaluate four base strategies detailed in Table~\ref{table:ablation}: \textbf{`Global (Original)`} using basic class names as a single global description; \textbf{`Global (GPT-3)`} using a single enriched action description generated by GPT-3; \textbf{`Static`} using predefined body-part partitions with multiple part-level descriptions; and our \textbf{`Adaptive`} strategy, where an attention mechanism dynamically identifies informative regions and learns to aggregate multi-granularity semantic representations.

The base model results show that richer semantics and finer-grained partitioning provide substantial gains. As indicated by the gains in parentheses, refinement consistently improves all models, but the magnitude depends on initial alignment quality. The `Adaptive` model achieves remarkably larger improvements from refinement compared to `Static` models, demonstrating a powerful synergistic effect between adaptive alignment and dynamic refinement.

\begin{table}[t]
\centering
\footnotesize
\caption{Ablation study (\%) on different alignment strategies and the effect of inference-time dynamic refinement. The `Dynamic` column indicates whether refinement is applied.}
\label{table:ablation}
\resizebox{\columnwidth}{!}{%
\begin{tabular}{lccccc}
\toprule
\multirow{2}{*}{\textbf{Strategy}} & \multirow{2}{*}{\textbf{Dynamic}} & \multicolumn{2}{c}{\textbf{NTU RGB+D 60}} & \multicolumn{2}{c}{\textbf{NTU RGB+D 120}} \\
\cmidrule(lr){3-4} \cmidrule(lr){5-6}
& & 55/5 & 48/12 & 110/10 & 96/24 \\
\midrule
\multirow{2}{*}{Global (Original)}
& & 64.69 & 35.46 & 66.96 & 44.27 \\
& \checkmark & 71.82 {\footnotesize (+7.13)} & 44.28 {\footnotesize (+8.82)} & 73.05 {\footnotesize (+6.09)} & 47.29 {\footnotesize (+3.02)} \\
\midrule
\multirow{2}{*}{Global (GPT-3)}
& & 78.50 & 33.47 & 64.89 & 47.15 \\
& \checkmark & 84.67 {\footnotesize (+6.17)} & 53.05 {\footnotesize (+19.58)} & 85.64 {\footnotesize (+20.75)} & 52.77 {\footnotesize (+5.62)} \\
\midrule
\multirow{2}{*}{Static}
& & 76.46 & 33.03 & 67.62 & 46.83 \\
& \checkmark & 78.53 {\footnotesize (+2.07)} & 36.80 {\footnotesize (+3.77)} & 70.25 {\footnotesize (+2.63)} & 48.08 {\footnotesize (+1.25)} \\
\midrule
\multirow{2}{*}{Adaptive}
& & 79.23 & 40.99 & 71.95 & 52.01 \\
& \checkmark & \textbf{88.52} {\footnotesize \textbf{(+9.29)}} & \textbf{71.80} {\footnotesize \textbf{(+30.81)}} & \textbf{89.06} {\footnotesize \textbf{(+17.11)}} & \textbf{69.11} {\footnotesize \textbf{(+17.10)}} \\
\bottomrule
\end{tabular}%
}
\end{table}

\noindent\textbf{Multi-Granularity Components.}
Table~\ref{table:ablation2} evaluates the contributions of body-part (BP) and temporal-interval (TI) alignment components. Both BP and TI components independently improve over the baseline, and their combination yields the best results. The BP component shows stronger synergy with refinement, as body-part descriptions (e.g., "left hand") are more concrete than abstract temporal descriptions. The substantial performance leap in all `Refinement` rows demonstrates that dynamic refinement compels the model to discover more discriminative alignment distributions.

\begin{table}[t]
\centering
\footnotesize
\caption{Ablation study (\%) analyzing the contribution of body-part (BP) and temporal-interval (TI) alignment with different aggregation strategies, with and without inference-time refinement.}
\label{table:ablation2}
\begin{tabular}{cllcccc}
\toprule
\multirow{2}{*}{\textbf{Method}} &
\multicolumn{1}{c}{\multirow{2}{*}{\textbf{BP}}} &
\multicolumn{1}{c}{\multirow{2}{*}{\textbf{TI}}} &
\multicolumn{2}{c}{\textbf{NTU RGB+D 60 }} &
\multicolumn{2}{c}{\textbf{NTU RGB+D 120 }} \\
& & & 55/5 & 48/12 & 110/10 & 96/24 \\ \midrule
Global-only & & & 78.50 & 33.47 & 64.89 & 47.15 \\ \midrule
Average & \checkmark & & 76.68 & 37.80 & 68.11 & 30.93 \\
Learnable & \checkmark & & 76.32 & 37.62 & 71.73 & 40.92 \\
Refinement & \checkmark & & 87.40 & 61.46 & 81.57 & 60.14 \\ \midrule
Average & & \checkmark & 78.65 & 38.80 & 55.73 & 50.67 \\
Learnable & & \checkmark & 77.70 & 40.69 & 71.26 & 46.13 \\
Refinement & & \checkmark & 87.25 & 55.42 & 80.40 & 58.74 \\ \midrule
Average & \checkmark & \checkmark & 79.02 & 39.92 & 73.55 & 51.38 \\
Learnable & \checkmark & \checkmark & 79.23 & 40.99 & 71.95 & 52.01 \\
Refinement & \checkmark & \checkmark & \textbf{88.52} & \textbf{71.80} & \textbf{89.06} & \textbf{69.11} \\
\bottomrule
\end{tabular}
\end{table}

\subsection{Dynamic Refinement Analysis}\label{subsec:refinement_analysis}

This subsection investigates the efficacy of the dynamic refinement mechanism, with a specific focus on the role of the memory bank and its impact on generalized zero-shot learning performance.

\noindent\textbf{Inference-Time Dynamic Refinement Framework.}
Table~\ref{tab:ablation_mb} evaluates three configurations: (1) \textbf{`Baseline`} without refinement; (2) \textbf{`w/o Memory Bank`} using naive pseudo-labeling; and (3) our \textbf{`Full Framework`} with class-balanced memory bank. While naive pseudo-labeling provides some benefit, our full framework delivers consistent and substantial improvements, including a 30.81\% gain on NTU-60 48/12, confirming the memory bank's critical role in stabilizing adaptation.

\begin{table}[t]
\centering
\footnotesize
\caption{Ablation study (\%) on test-time refinement components. Results show ZSL accuracy and GZSL harmonic mean.}
\label{tab:ablation_mb}
\resizebox{\columnwidth}{!}{%
\begin{tabular}{lcccccccc}
\toprule
\multirow{2}{*}{\textbf{Method}} &
\multicolumn{4}{c}{\textbf{NTU RGB+D 60}} &
\multicolumn{4}{c}{\textbf{NTU RGB+D 120}} \\
& \multicolumn{2}{c}{55/5} & \multicolumn{2}{c}{48/12} & \multicolumn{2}{c}{110/10} & \multicolumn{2}{c}{96/24} \\
& ZSL & GZSL & ZSL & GZSL & ZSL & GZSL & ZSL & GZSL \\ \midrule
Baseline & 79.22 & 64.35 & 40.99 & 48.74 & 71.95 & 72.97 & 52.01 & 53.99 \\
w/o Memory Bank & 78.02 & 63.83 & 51.60 & 56.99 & 81.20 & 77.09 & 59.82 & 56.95 \\
Full Framework & \textbf{88.52} & \textbf{69.95} & \textbf{71.80} & \textbf{70.66} & \textbf{89.06} & \textbf{81.49} & \textbf{69.11} & \textbf{64.36} \\
\bottomrule
\end{tabular}%
}
\end{table}

\noindent\textbf{Domain-Specific Refinement in GZSL.}
To understand the impact of our test-time refinement in the GZSL setting, we analyze its effect on seen (S) and unseen (U) classes separately. Table~\ref{tab:ablation_tta} details the performance of the \textbf{`Baseline`} model (no refinement), a \textbf{`Partial`} model adapted only on seen class data, and our \textbf{`Full`} approach that adapts on the entire test distribution.

The results reveal an important and desirable asymmetry. Our `Full` refinement strategy primarily benefits the unseen classes, with accuracy gains of up to 24.89\%, while performance on seen classes is preserved. This outcome supports our motivation that unseen classes suffer more from semantic drift and thus benefit most from refinement. Consequently, the harmonic mean (H) improves dramatically by up to 19.82\%, demonstrating that our approach successfully mitigates bias towards seen classes and achieves a much better performance balance across both domains.

\begin{table}[t]
\centering
\footnotesize
\caption{Analysis of test-time refinement impact on seen (S) and unseen (U) classes in GZSL, with harmonic mean (H).}
\label{tab:ablation_tta}
\resizebox{\columnwidth}{!}{%
\begin{tabular}{lcccccccccccc}
\toprule
\multirow{2}{*}{\textbf{Method}} &
\multicolumn{6}{c}{\textbf{NTU RGB+D 60}} &
\multicolumn{6}{c}{\textbf{NTU RGB+D 120}} \\
& \multicolumn{3}{c}{55/5} & \multicolumn{3}{c}{48/12} & \multicolumn{3}{c}{110/10} & \multicolumn{3}{c}{96/24} \\
& S & U & H & S & U & H & S & U & H & S & U & H \\ \midrule
Baseline & 71.70 & 60.35 & 65.53 & 81.60 & 36.92 & 50.84 & 76.25 & 67.89 & 71.80 & 72.67 & 45.32 & 55.81 \\
Partial & 72.67 & 60.35 & 65.93 & 82.48 & 36.92 & 51.01 & 80.02 & 67.89 & 73.46 & 74.41 & 45.32 & 56.33 \\
Full & \textbf{72.67} & \textbf{67.43} & \textbf{69.95} & \textbf{82.48} & \textbf{61.81} & \textbf{70.66} & \textbf{80.02} & \textbf{83.00} & \textbf{81.49} & \textbf{74.41} & \textbf{56.70} & \textbf{64.36} \\
\bottomrule
\end{tabular}%
}
\end{table}

\begin{table}[t]
\centering
\footnotesize
\caption{Comparison with other TTA methods on NTU RGB+D 60 and NTU RGB+D 120 datasets.}
\label{tab:tta-comparison}
\resizebox{0.9\columnwidth}{!}{%
\begin{tabular}{@{}lccccc@{}}
\toprule
\multirow{2}{*}{\textbf{Method}} & \multicolumn{2}{c}{\textbf{NTU RGB+D 60}} & \multicolumn{2}{c}{\textbf{NTU RGB+D 120}} & \multirow{2}{*}{\textbf{Throughput}} \\
\cmidrule(lr){2-3} \cmidrule(lr){4-5}
& \textbf{55/5} & \textbf{48/12} & \textbf{110/10} & \textbf{96/24} & \textbf{(samples/ms)} \\ \midrule
TPT \cite{shu2022test} & 40.11 & 27.98 & 22.10 & 17.10 & 0.70 \\
AdaNPC \cite{zhang2023adanpc} & 81.77 & 42.52 & 72.50 & 53.16 & 2.15  \\
CALIP~\cite{guo2022calipzeroshotenhancementclip} & 80.02 & 45.10 & 72.27 & 51.84 & 2.48 \\
TDA~\cite{karmanov2024efficient} & 82.84 & 43.95 & 74.92 & 54.60 & 3.42 \\
\midrule
PURLS (baseline) & 79.22 & 40.99 & 71.95 & 52.01 & 10.09 \\
\textbf{DynaPURLS (Ours)} & \textbf{88.52} & \textbf{71.80} & \textbf{89.06} & \textbf{69.11} & 1.95 \\
\bottomrule
\end{tabular}%
}
\end{table}

\noindent\textbf{Comparison with Alternative Adaptation Methods.}
To contextualize our contribution, we compare our dynamic refinement strategy against existing state-of-the-art TTA methods. For a fair comparison of the adaptation strategies themselves, all methods are applied to a frozen \textbf{PURLS} backbone. As shown in Table~\ref{tab:tta-comparison}, our \textbf{DynaPURLS} achieves substantial improvements across all evaluation scenarios, with accuracy gains of 5.68\% and 27.85\% over the strongest baseline (TDA) on NTU RGB+D 60, and 14.14\% and 14.51\% on NTU RGB+D 120, demonstrating consistent superior performance.

Existing baselines exhibit significant limitations when applied to skeleton data. Gradient-based prompt tuning methods (TPT~\cite{shu2022test}) require iterative optimization that easily overfits the small and highly imbalanced target streams characteristic of skeleton action recognition. Prototype and attention-based schemes (AdaNPC~\cite{zhang2023adanpc}, CALIP~\cite{guo2022calipzeroshotenhancementclip}) compress each clip into a single holistic vector, discarding fine-grained joint dynamics and thereby conflating actions that differ only in local motion patterns. TDA~\cite{karmanov2024efficient} partially addresses these issues with a dual cache mechanism, yet it still operates at a single global scale and ignores temporal phase diversity, limiting its effectiveness on complex multi-phase actions. In contrast, \textbf{DynaPURLS} introduces dynamic, confidence-guided alignment: a lightweight transformation refines textual features at multiple granularities rather than compressing features or updating the backbone directly, thereby preserving both body-part specificity and temporal phase information. This refinement is further stabilized by a class-balanced memory bank that prevents bias toward dominant classes and ensures robust adaptation over imbalanced data streams, yielding the superior empirical performance shown in the table.

\subsection{Hyperparameter and Initialization Studies}\label{subsec:hyperparameter_initialization}

This subsection examines the robustness of the proposed approach through sensitivity analysis of key hyperparameters and an evaluation of initialization strategies.

\begin{figure}[t]
\centering
\includegraphics[width=0.45\textwidth]{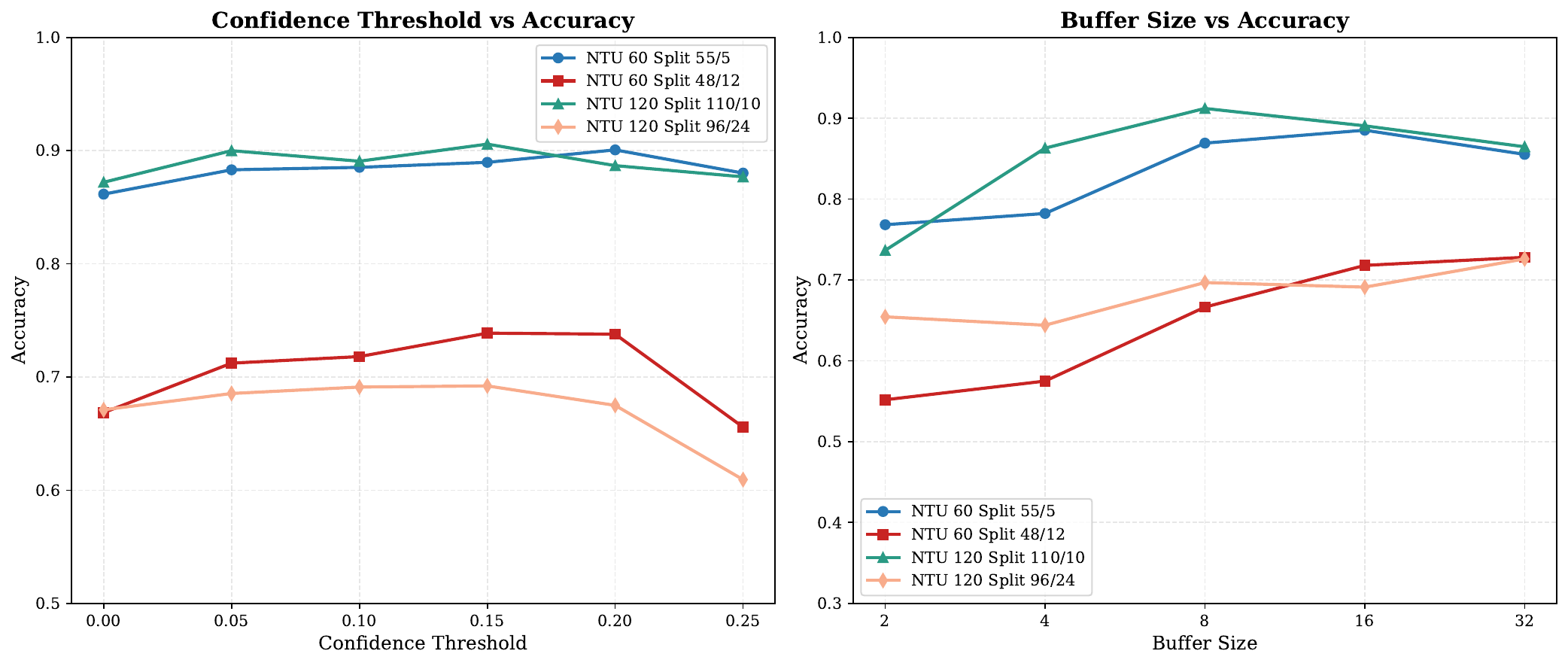}
\caption{Hyperparameter sensitivity analysis on NTU RGB+D dataset. (a) Effect of confidence threshold $\tau$ on refinement quality. (b) Impact of memory bank capacity K on performance.}
\label{fig:hyperparameters}
\vspace{-0.2cm}
\end{figure}

\noindent\textbf{Key Hyperparameters of Refinement Framework.} We analyze the model's sensitivity to key hyperparameters of our refinement framework. Figure~\ref{fig:hyperparameters} presents the performance analysis on the NTU RGB+D dataset, revealing important trade-offs for optimal performance.

\noindent\textbf{Confidence Threshold.} The confidence threshold $\tau$ (panel a) acts as an initial filter for noisy samples. Performance improves as $\tau$ increases to 0.2 by selecting higher-quality samples, but declines sharply beyond this as misclassified samples dominate the refinement process. A lower initial $\tau$ is preferable, as our memory bank iteratively improves sample quality over time; an overly restrictive threshold hinders refinement and risks overfitting.

\noindent\textbf{Memory Bank Capacity.} The memory bank capacity $K$ (panel b) introduces a critical trade-off. An overly small $K$ limits sample diversity and risks overfitting, while an excessively large $K$ can be compromised by samples with incorrect pseudo-labels. This is reflected in the performance, which first increases and then declines as $K$ grows. Our empirical results suggest that a capacity of $K \in [8, 16]$ strikes an optimal balance.

These results confirm that while our method is robust within reasonable parameter ranges, understanding these trade-offs is key to maximizing its performance.

\noindent\textbf{Initialization Strategy.} To investigate the robustness of our dynamic refinement module to different initialization strategies, we compare three initialization schemes for the adaptive parameters $\mathbf{S}$ (scaling tensor) and $\Delta\mathbf{F}$ (bias tensor): (1) \textbf{Identity}: $\mathbf{S} = \mathbf{1}$ and $\Delta\mathbf{F} = \mathbf{0}$, which starts with no transformation; (2) \textbf{Random}: uniformly random initialization in $[-0.1, 0.1]$; and (3) \textbf{Kaiming}: Kaiming uniform initialization~\cite{he2015delving}. Table~\ref{tab:initialization} presents the results across multiple dataset splits.

\begin{table}[t]
\centering
\footnotesize
\caption{Ablation study (\%) on initialization strategies for adaptive parameters. Results show ZSL accuracy across different dataset splits.}
\label{tab:initialization}
\begin{tabular}{lcccc}
\toprule
\multirow{2}{*}{\textbf{Initialization}} & \multicolumn{2}{c}{\textbf{NTU RGB+D 60}} & \multicolumn{2}{c}{\textbf{NTU RGB+D 120}} \\
\cmidrule(lr){2-3} \cmidrule(lr){4-5}
& 55/5 & 48/12 & 110/10 & 96/24 \\
\midrule
Identity ($\mathbf{S}=\mathbf{1}$, $\Delta\mathbf{F}=\mathbf{0}$) & \textbf{88.52} & \textbf{71.80} & \textbf{89.06} & \textbf{69.11} \\
Random & 88.31 & 71.45 & 88.89 & 68.87 \\
Kaiming & 88.44 & 71.63 & 88.95 & 69.02 \\
\bottomrule
\end{tabular}
\end{table}

The results reveal that our method is remarkably robust to initialization choices, with performance variations of less than 0.5\% across all splits. The identity initialization slightly outperforms alternatives, which is intuitive as it begins from the well-trained static semantic embeddings and allows the refinement process to gradually discover beneficial adaptations guided by test data. Random and Kaiming initializations introduce initial perturbations that must be corrected during the early refinement steps, resulting in marginally lower final performance. Nevertheless, the negligible differences confirm that our confidence-guided optimization with class-balanced memory bank provides a stable refinement trajectory regardless of initialization, highlighting the robustness of our approach.

\subsection{Qualitative Analysis}\label{subsec:qualitative}

To provide a qualitative perspective on the model's improvements, we present visualizations of class-wise performance changes, confusion matrices, and feature distribution evolution.

\begin{figure}[tb]
    \centering
    \includegraphics[width=0.75\linewidth]{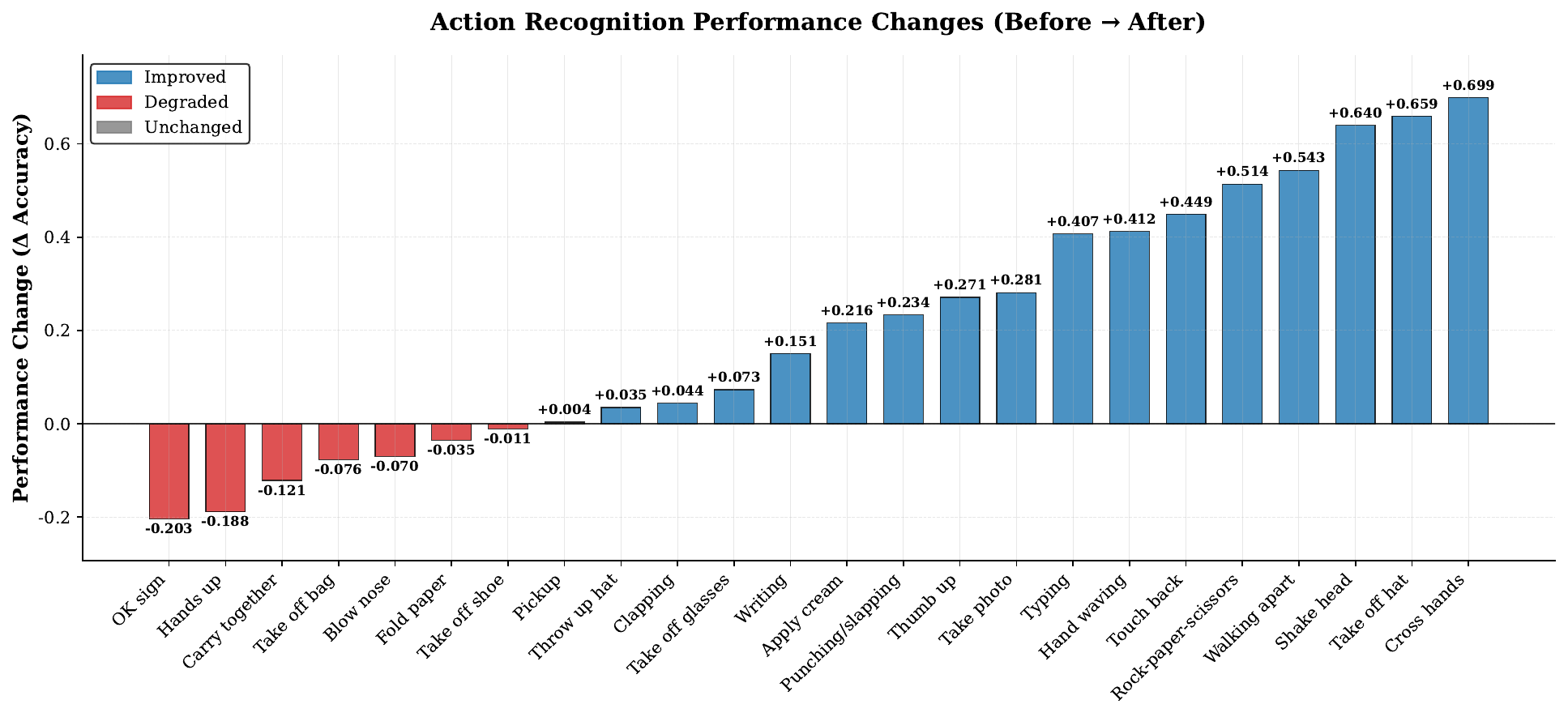}
    \caption{Class-wise performance changes from PURLS to DynaPURLS on NTU RGB+D 120 (96/24 split). Actions are sorted by performance change, with blue bars indicating improvement, red bars indicating degradation, and gray indicating negligible change.}
    \label{fig:classwise_performance}
    \vspace{-0.3cm}
\end{figure}

\noindent\textbf{Class-wise Performance Patterns.} Figure~\ref{fig:classwise_performance} presents a detailed class-wise comparison between PURLS and DynaPURLS on the NTU RGB+D 120 dataset (96/24 split). The most significant improvements occur in actions with distinctive motion patterns: ``Cross hands'' (+0.699), ``Take off hat'' (+0.659), and ``Shake head'' (+0.640). Conversely, actions like ``OK sign'' (-0.204) and ``Hands up'' (-0.188) show performance degradation due to visual ambiguity with similar gestures. Overall, DynaPURLS improves performance on 17 out of 24 unseen classes (70.8\%), demonstrating that our test-time refinement is particularly effective for actions with unambiguous motion patterns.

\begin{figure}[tb]
    \centering
    \footnotesize
    \subfloat[PURLS]{\includegraphics[width=0.38\linewidth]{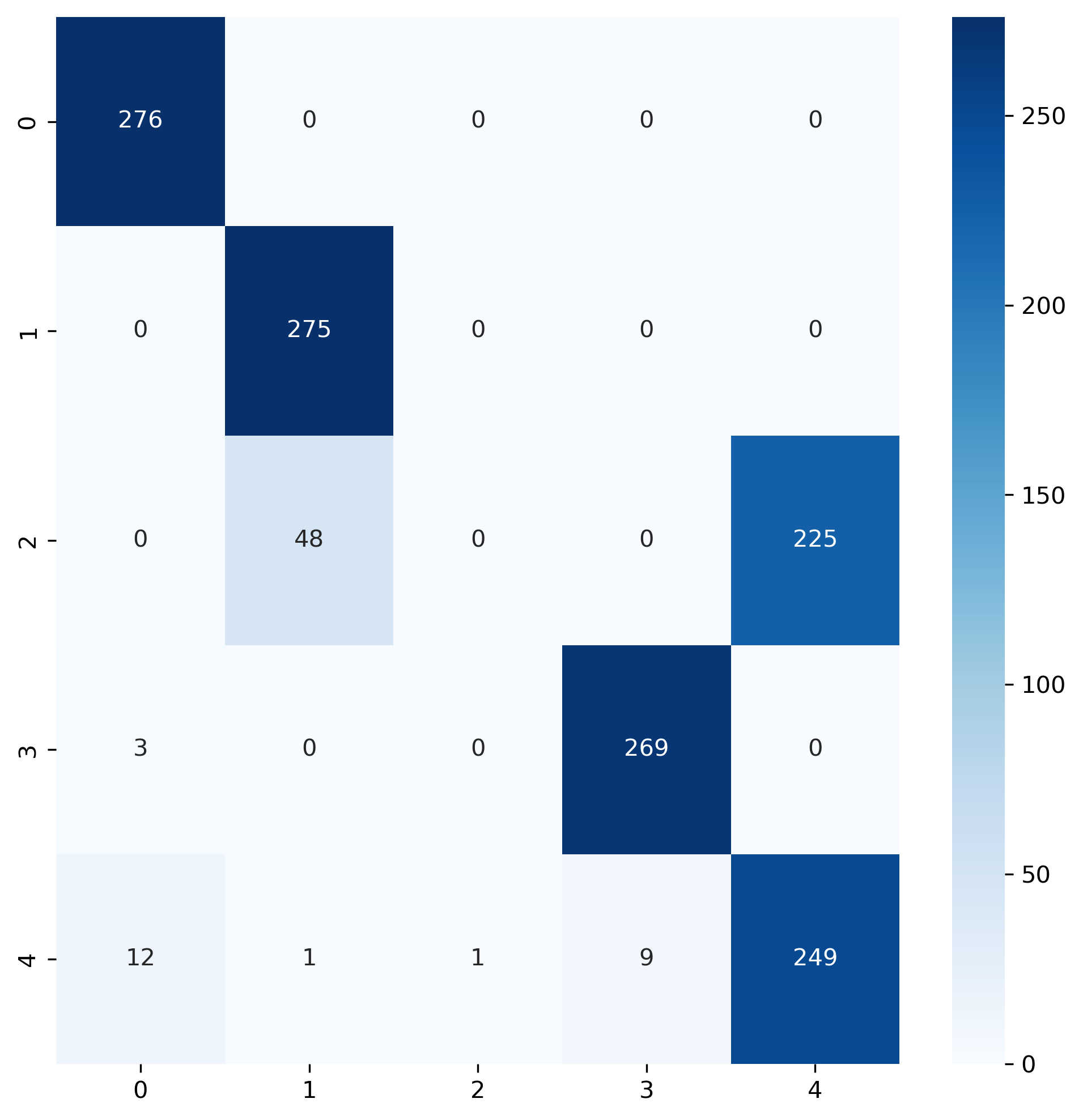}}
    \quad
    \subfloat[DynaPURLS]{\includegraphics[width=0.38\linewidth]{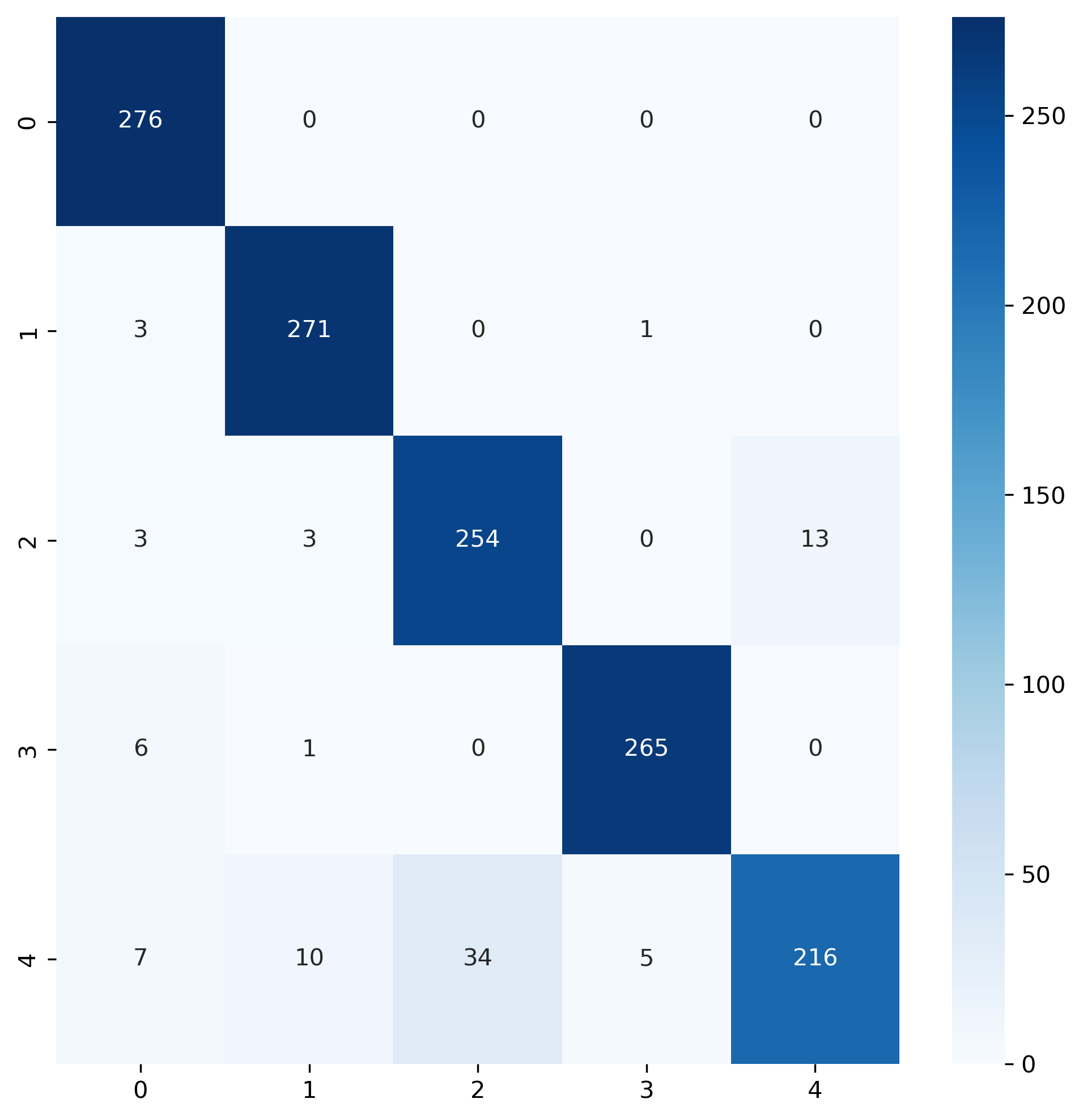}}
    \caption{Confusion matrices comparison for NTU RGB+D 55/5 split before and after refinement with DynaPURLS.}
    \label{fig:confusion_matrices}
    \vspace{-0.2cm}
\end{figure}

\noindent\textbf{Confusion Matrix Visualization.} Figure~\ref{fig:confusion_matrices} compares confusion matrices before and after refinement with DynaPURLS for the NTU RGB+D 55/5 split. Before refinement, the model struggles to distinguish similar actions like \textit{reading} and \textit{writing}. Post-refinement, the diagonal sharpens and off-diagonal elements lighten, indicating improved class differentiation.

\begin{figure}[t]
    \centering
    \footnotesize
    \begin{subfigure}[b]{0.38\linewidth}
        \centering
        \includegraphics[width=0.9\textwidth, trim=100 100 100 100, clip]{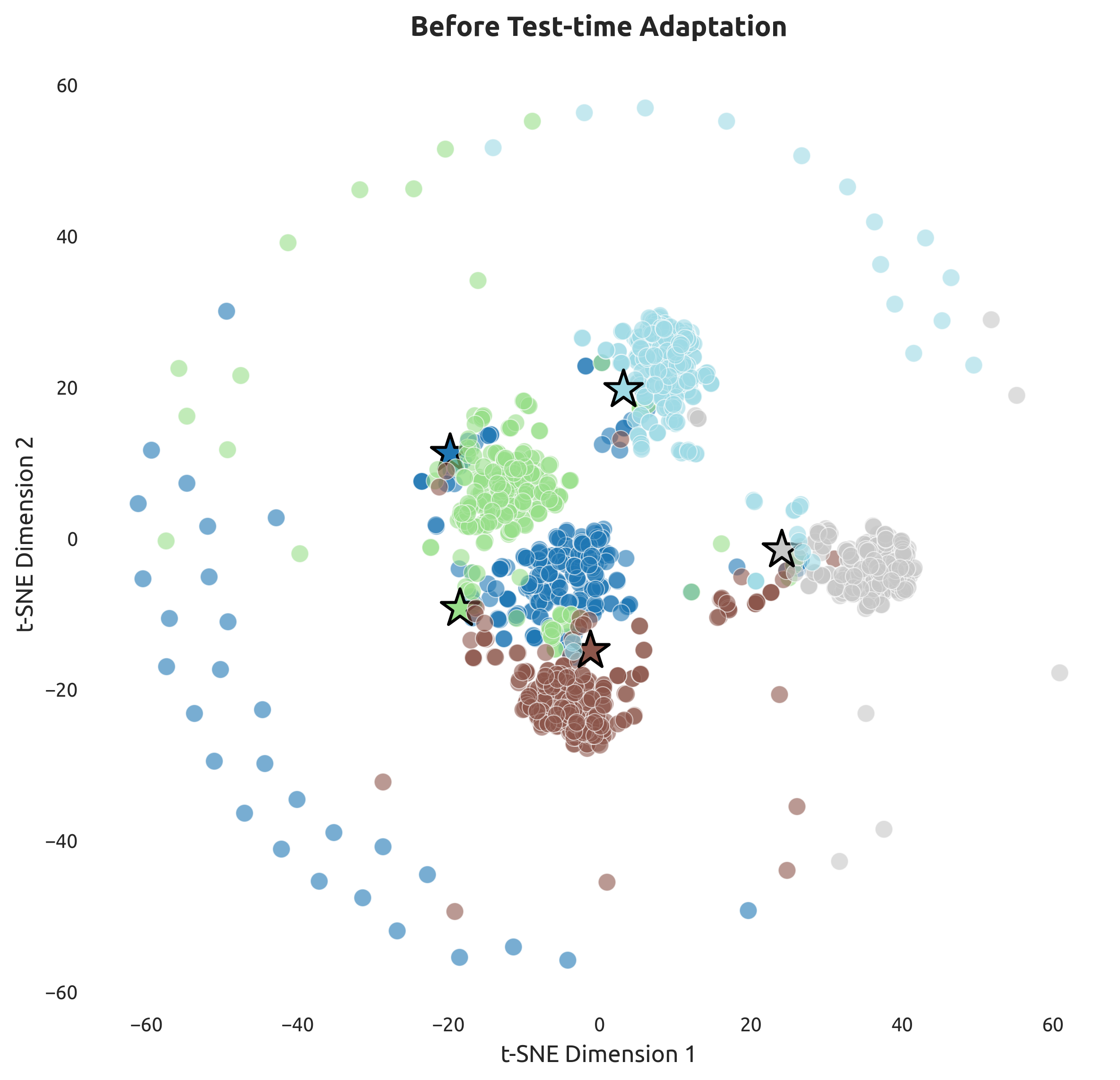}
        \caption{Stage 1}
    \end{subfigure}
    \hfill
    \begin{subfigure}[b]{0.38\linewidth}
        \centering
        \includegraphics[width=0.9\textwidth, trim=100 100 100 100, clip]{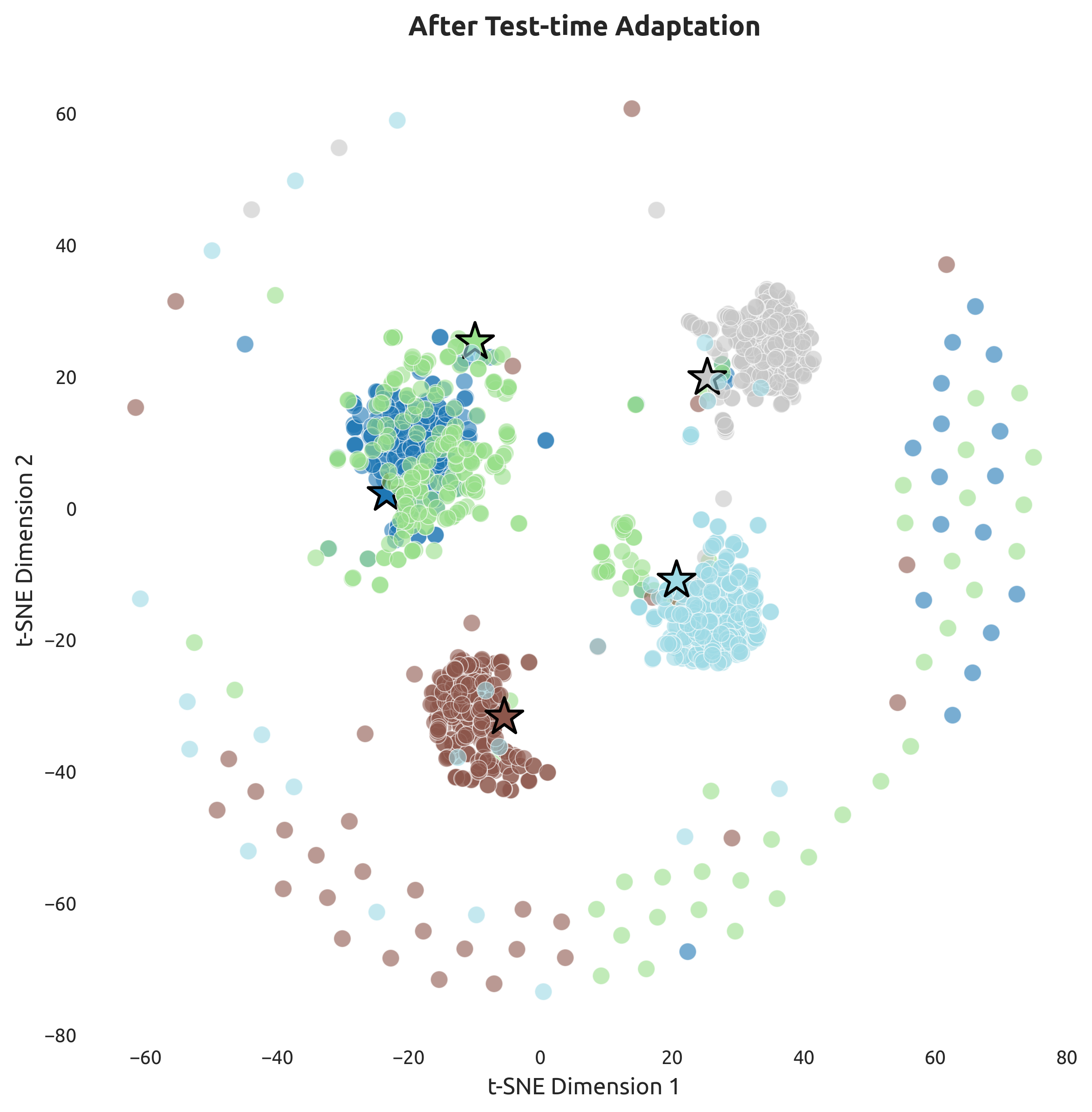}
        \caption{Stage 2}
    \end{subfigure}
    \vspace{0.5mm}
    \begin{subfigure}[b]{0.38\linewidth}
        \centering
        \includegraphics[width=\textwidth, trim=100 100 100 100, clip]{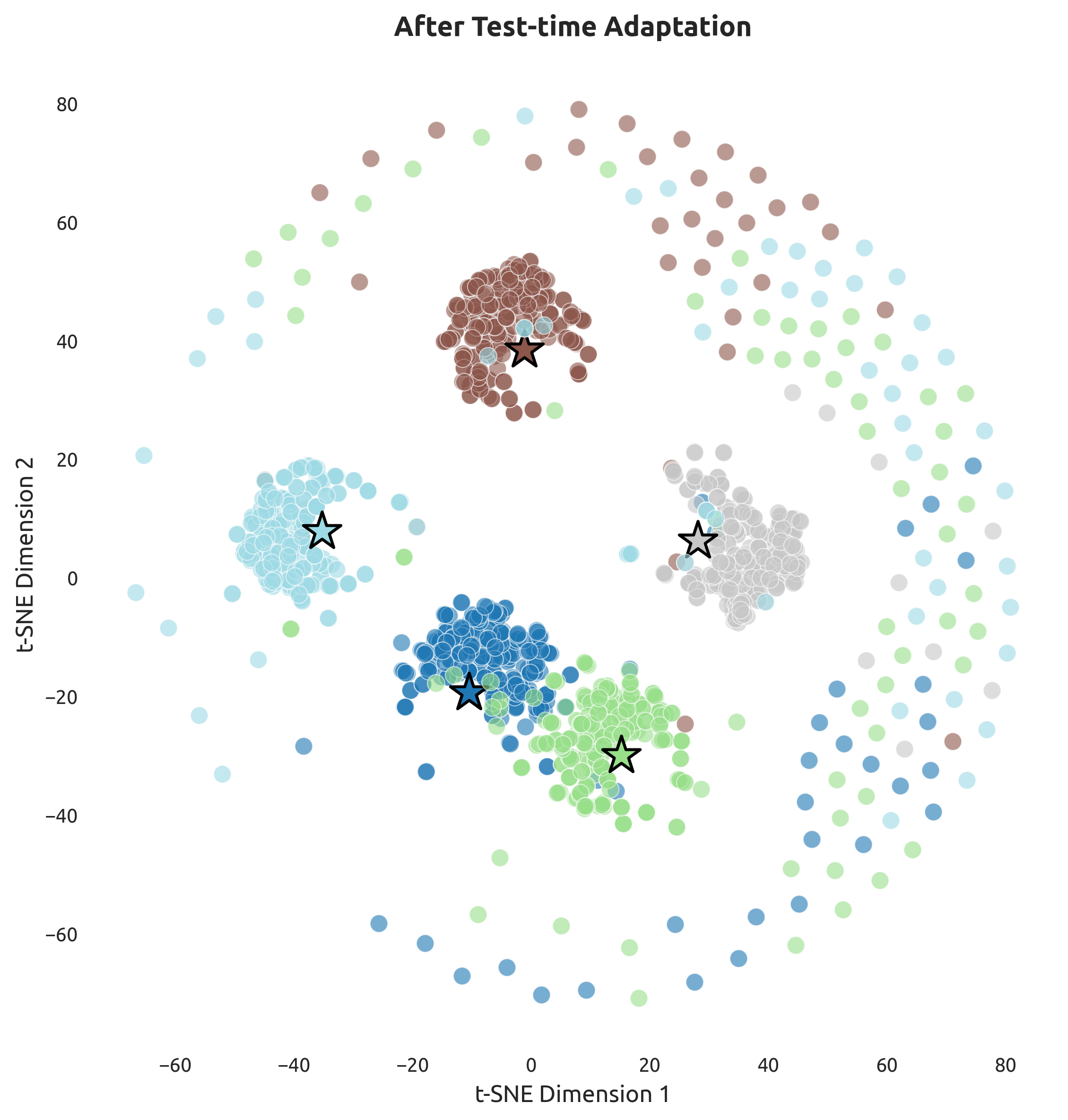}
        \caption{Stage 3}
    \end{subfigure}
    \hfill
    \begin{subfigure}[b]{0.38\linewidth}
        \centering
        \includegraphics[width=\textwidth, trim=100 100 100 100, clip]{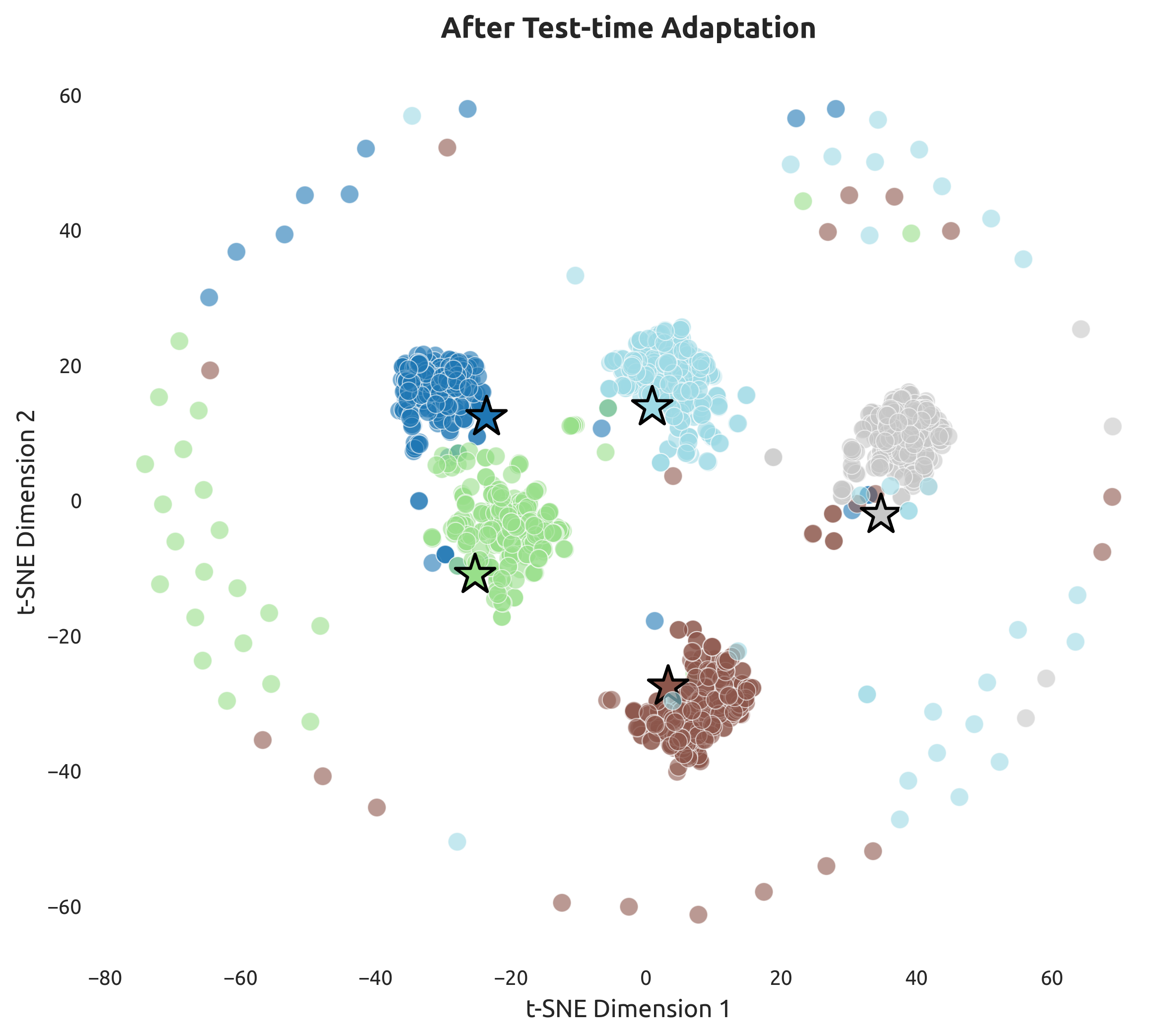}
        \caption{Stage 4}
    \end{subfigure}
    \caption{t-SNE visualization of feature distributions during progressive refinement on NTU RGB+D 60, where $\star$ represents text features and $\bullet$ represents skeleton features, with different colors denoting distinct classes.}
    \label{fig:tsne2}
    \vspace{-0.2cm}
\end{figure}

\noindent\textbf{Progressive Refinement Stages.}
To evaluate the impact of progressive refinement, we visualize feature distributions using t-SNE, dividing the test set into four parts corresponding to each stage. As shown in Fig.~\ref{fig:tsne2}, visual features from different classes exhibit significant overlap at Stage 1. With test-time refinement, Stage 2 achieves more distinct class boundaries. In Stage 3, further refinement reduces overlap, and by Stage 4, the overlapping issue is largely resolved, demonstrating the effectiveness of our progressive refinement strategy.

\section{Discussion}

While DynaPURLS achieves state-of-the-art performance, our analysis reveals several limitations providing insights for future research.

\noindent\textbf{Fine-grained Action Disambiguation.} Our method exhibits reduced performance on action pairs sharing highly similar motion patterns. As shown in Figure~\ref{fig:confusion_matrices}, distinguishing ``writing'' from ``drawing'' remains challenging, as both involve similar hand trajectories differing primarily in subtle wrist movements. Future work could explore adversarial training between confusing action pairs or attention mechanisms emphasizing discriminative spatial-temporal regions.

\noindent\textbf{Dataset Biases and Limited Generalization.} Current evaluation is confined to controlled indoor environments. NTU RGB+D~\cite{ntu60,ntu120} and PKU-MMD~\cite{liu2017pku} predominantly feature young adults performing scripted actions in laboratory settings, which may not reflect natural movement variability across cultures, age groups, and contexts. Addressing this requires evaluation on diverse datasets and domain adaptation techniques bridging controlled and naturalistic settings.

\noindent\textbf{Memory Bank Reliability and Semantic Drift.} While our class-balanced memory bank stabilizes refinement, it assumes high-confidence predictions correlate with correctness. However, the model can be confidently wrong, particularly for actions visually similar to well-represented seen classes. Future methods could benefit from sophisticated uncertainty estimation techniques to identify truly reliable samples.

\noindent\textbf{Temporal Dynamics and Sequential Dependencies.} Our approach treats temporal segments relatively independently through simple start-middle-end decomposition, which may not adequately capture complex temporal dynamics. Incorporating variable-length temporal segmentation or hierarchical representations could better model sequential movement patterns.

These limitations highlight the challenging nature of zero-shot skeleton-based action recognition and suggest important directions for future research.

\section{Conclusion}

In this paper, we introduced DynaPURLS, a novel framework that advances zero-shot skeleton-based action recognition through multi-granularity semantic alignment and dynamic test-time refinement. Our approach leverages LLM-generated hierarchical descriptions and adaptive cross-modal attention to establish fine-grained visual-semantic correspondences. To overcome the limitations of static representations, we introduced the first test-time adaptation mechanism specifically designed for skeleton-based zero-shot learning, which dynamically refines semantic embeddings via confidence-guided optimization stabilized by a class-balanced memory bank. Extensive experiments on NTU RGB+D 60/120 and PKU-MMD demonstrate substantial improvements over state-of-the-art methods in both ZSL and GZSL settings.

\section*{Acknowledgment}
This research was supported by the Australian Government through the Australian Research Council's DECRA funding scheme (Grant No.: DE250100030) and Discovery Project funding scheme (Grant No.: DP260100218, DP260101891). Dr Qiuhong Ke is the recipient of an Australian Research Council Discovery Early Career Researcher Award (project number DE250100030) funded by the Australian Government.

\bibliographystyle{IEEEtran}
\bibliography{IEEEabrv,reference} %

\def\BioVSkip{\vspace{-20pt}}

\begin{IEEEbiography}[{\includegraphics[width=0.85in,height=1.05in,clip,keepaspectratio]{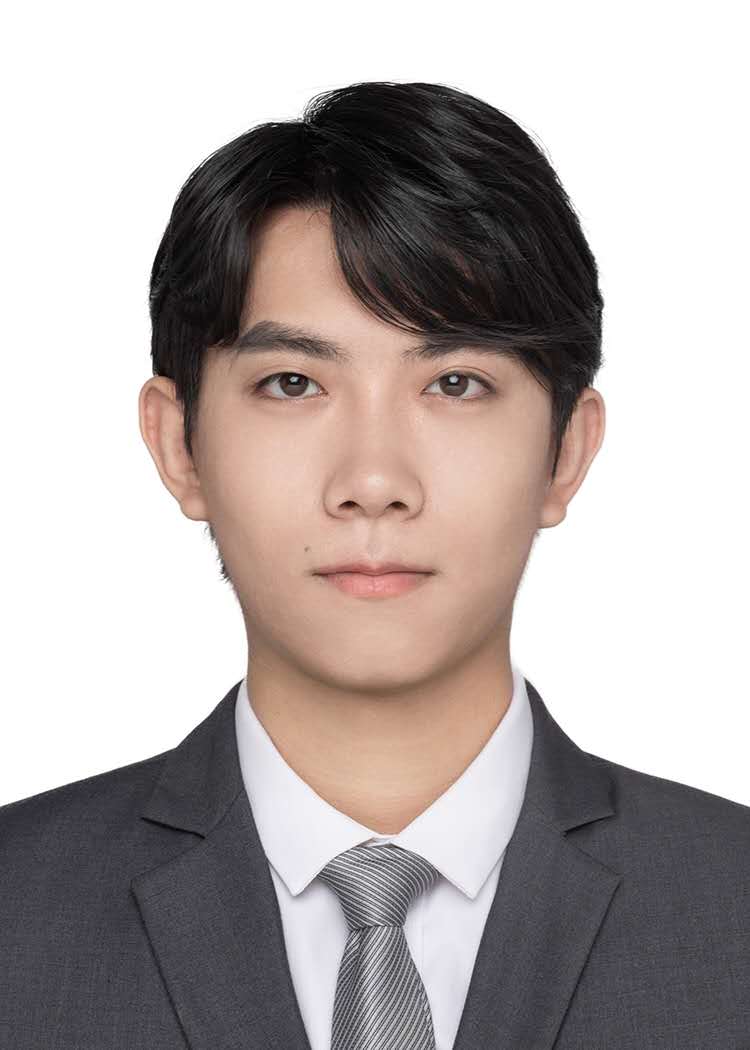}}]{Jingmin Zhu}
received his Bachelor's degree in Physics from Southern University of Science and Technology, Shenzhen, China, in July 2023, and his Master's degree in Artificial Intelligence from Monash University, Melbourne, Australia, in July 2025. His current research interests include computer vision, multimodal large language models, domain adaptation, zero-shot learning, and few/one-shot learning.
\end{IEEEbiography}\BioVSkip

\begin{IEEEbiography}[{\includegraphics[width=0.85in,height=1.05in,clip,keepaspectratio]{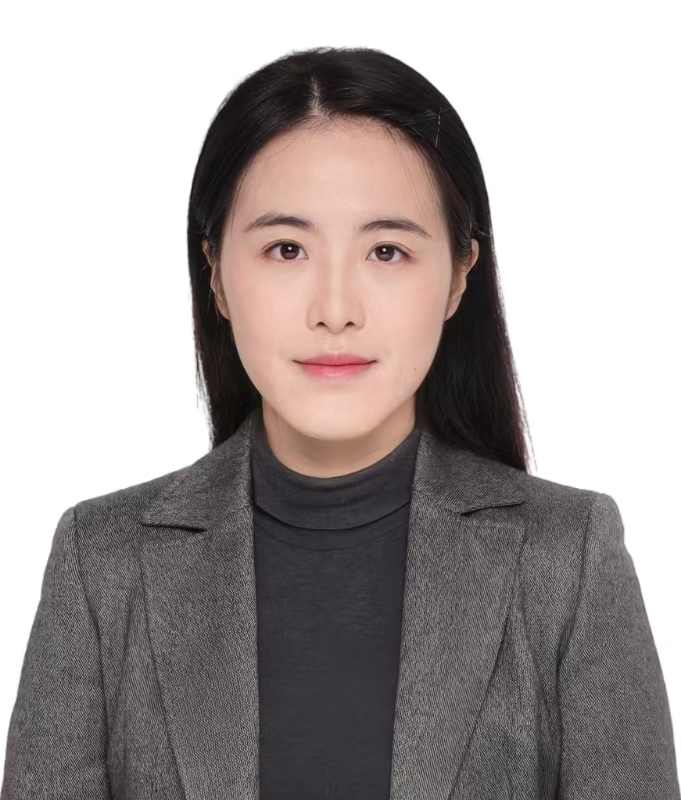}}]{Anqi Zhu}
received her Bachelor's degree in Software Engineering from the University of New South Wales, Sydney, Australia, in 2021, and her Ph.D. degree in Engineering \& IT from the University of Melbourne, Melbourne, Australia, in 2025. She subsequently joined Monash University, Melbourne, Australia, as a researcher in February 2025. Her current research interests include computer vision, skeleton-based action recognition, zero-shot learning, and few/one-shot learning.
\end{IEEEbiography}\BioVSkip

\begin{IEEEbiography}[{\includegraphics[width=0.85in,height=1.05in,clip,keepaspectratio]{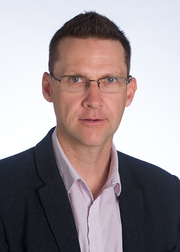}}]{James Bailey}
is a Professor and Head of the Department of Data Science and Artificial Intelligence in the Faculty of Information Technology at Monash University. He was previously an Australian Research Council Future Fellow and is a researcher in the field of machine learning and artificial intelligence, including interdisciplinary applications and operational frameworks. His interests particularly relate to the assurance, certification and safety of systems based on machine learning and artificial intelligence. He works on the deployment of AI systems in collaboration with a wide range of industry and government partners across the defence, energy and health sectors.
\end{IEEEbiography}\BioVSkip

\begin{IEEEbiography}[{\includegraphics[width=0.85in,height=1.05in,clip,keepaspectratio]{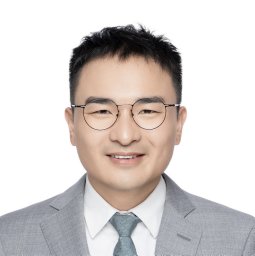}}]{Jun Liu}
is a Professor at School of Computing and Communications in Lancaster University. He got the PhD degree from Nanyang Technological University in 2019. He was with Singapore University of Technology and Design from 2019 to 2024. He is a Senior Area Editor of IEEE Transactions on Image Processing, and an Associate Editor of IEEE Transactions on Circuits and Systems for Video Technology, IEEE Transactions on Neural Networks and Learning Systems, IEEE Transactions on Industrial Informatics, ACM Computing Surveys, and Pattern Recognition. He is General Chair of BMVC 2026 and Program Chair of BMVC 2025. He has served as an Area Chair of CVPR, ECCV, ICML, NeurIPS, ICLR, IJCAI, AAAI, WACV, and MM. His research interests include computer vision, machine learning and digital health.
\end{IEEEbiography}\BioVSkip

\begin{IEEEbiography}[{\includegraphics[width=0.85in,height=1.05in,clip,keepaspectratio]{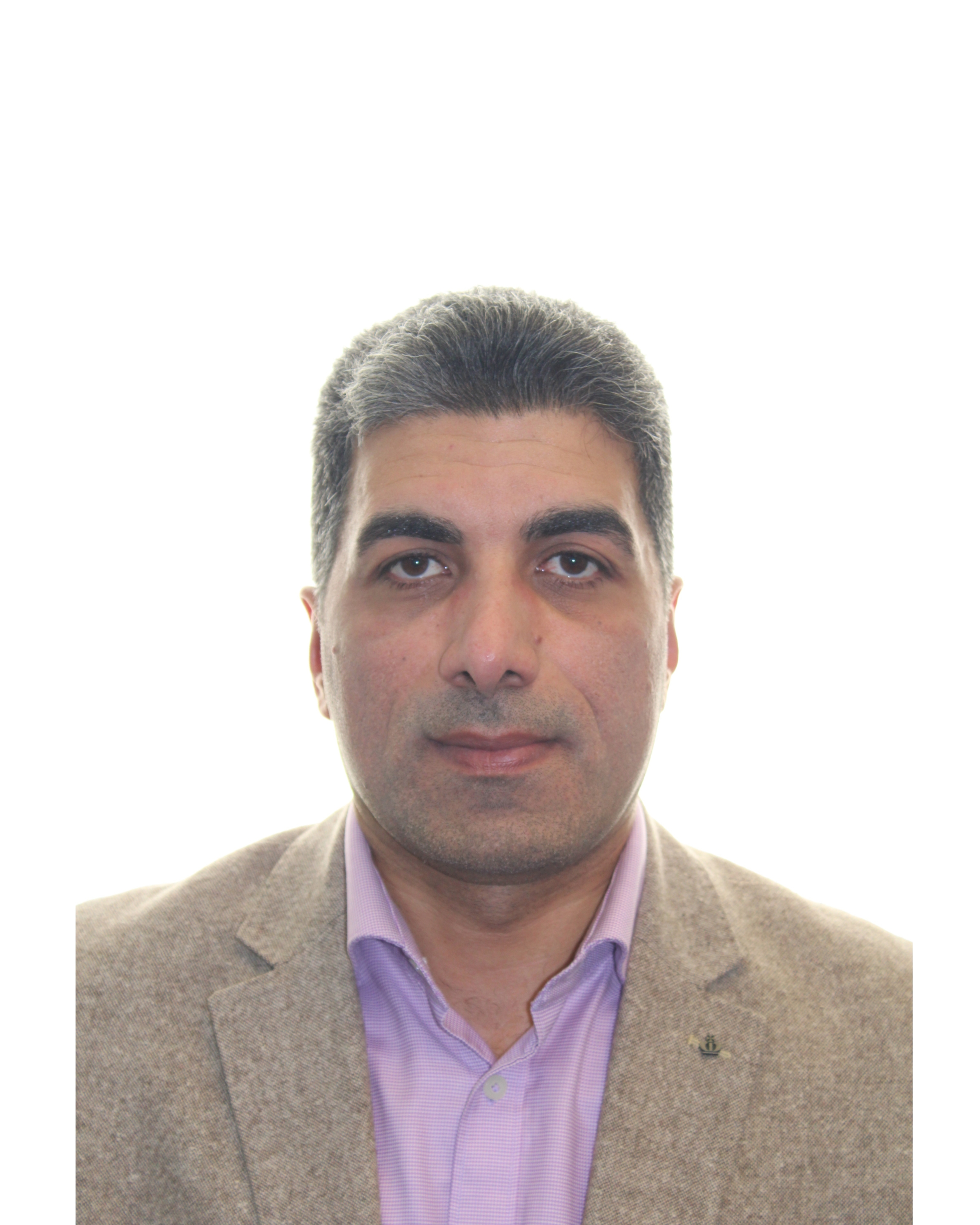}}]{Hossein Rahmani}
is a Professor in the School of Computing and Communications at Lancaster University, U.K. His research focuses on computer vision, machine learning, and generative AI. He has extensive leadership and administrative experience in the academic community. He currently serves as an Associate Editor for IEEE Transactions on Neural Networks and Learning Systems, Pattern Recognition, and ACM Computing Surveys. He serves as an Area Chair for major conferences including CVPR, ECCV, NeurIPS, and ICLR, and is a Program Chair of BMVC 2026. He has been awarded several competitive research grants supporting projects in computer vision and deep learning.
\end{IEEEbiography}\BioVSkip

\begin{IEEEbiography}[{\includegraphics[width=0.85in,height=1.05in,clip,keepaspectratio]{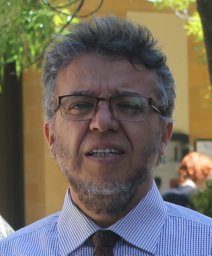}}]{Mohammed Bennamoun}
is Winthrop Professor in the Department of Computer Science and Software Engineering at UWA and is a researcher in computer vision, machine/deep learning, robotics, and signal/speech processing. He has published 4 books (available on Amazon), 1 edited book, 1 Encyclopedia article, 14 book chapters, 240+ journal papers, 300+ conference publications, 16 invited \& keynote publications. His h-index is 86 and his number of citations is 41,000+ (Google Scholar). He was awarded 65+ competitive research grants, from the Australian Research Council, and numerous other Government, UWA and industry Research Grants. He successfully supervised 45+ PhD students to completion. He won the Best Supervisor of the Year Award at QUT (1998), and received award for research supervision at UWA (2008 \& 2016) and Vice-Chancellor Award for mentorship (2016). He is also a Clarivate Highly Cited Researcher for 2025 (ranking among the top 0.1\% of scientists worldwide) and was recognised in The Australian's 2026 list of Australia's top 250 researchers, which highlights leading scholars across 250 research fields. He delivered conference tutorials at major conferences, including: IEEE Computer Vision and Pattern Recognition (CVPR 2016), Interspeech 2014, IEEE International Conference on Acoustics Speech and Signal Processing (ICASSP) and European Conference on Computer Vision (ECCV). He was also invited to give a Tutorial at an International Summer School on Deep Learning (DeepLearn 2017). He received Outstanding Paper Awards for his publications in \emph{IEEE Computational Intelligence Magazine} and the \emph{Proceedings of the IEEE}.
\end{IEEEbiography}\BioVSkip

\begin{IEEEbiography}[{\includegraphics[width=0.85in,height=1.05in,clip,keepaspectratio]{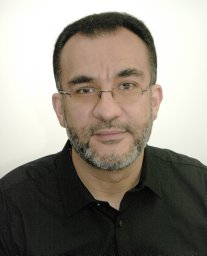}}]{Farid Boussaid}
received the M.S. and Ph.D. degrees in microelectronics from the National Institute of Applied Science, Toulouse, France, in 1996 and 1999, respectively. He joined Edith Cowan University, Perth, Australia, as a Post-Doctoral Research Fellow; and the Visual Information Processing Research Group, as a member, in 2000. He joined The University of Western Australia, Crawley, Australia, in 2005, where he is currently a Professor. His current research interests include smart CMOS sensors, computer vision, and machine learning.
\end{IEEEbiography}\BioVSkip

\begin{IEEEbiography}[{\includegraphics[width=0.85in,height=1.05in,clip,keepaspectratio]{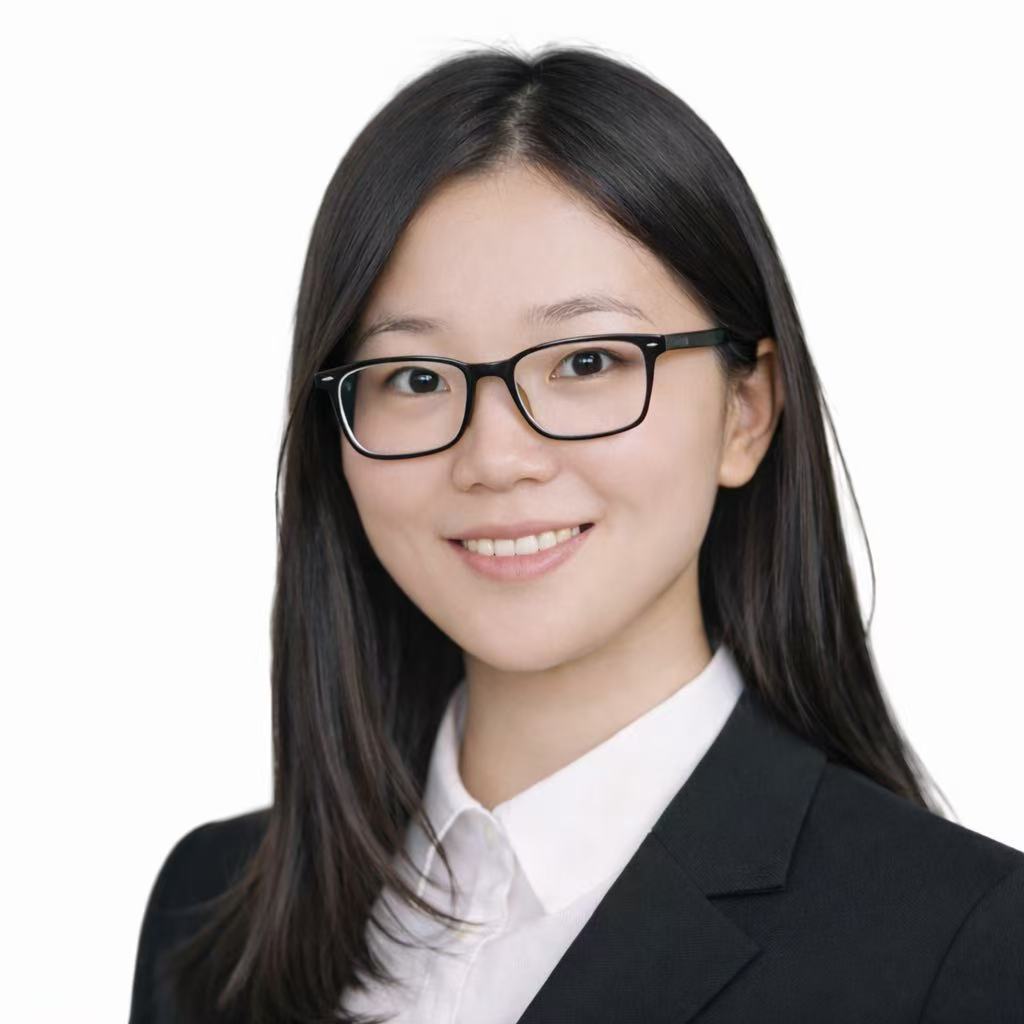}}]{Qiuhong Ke}
is an ARC DECRA Fellow and a Senior Lecturer at the Faculty of Information Technology, Monash University. Previously, she was a Postdoctoral Researcher at Max Planck Institute for Informatics from 2018 to 2019 and a Lecturer at The University of Melbourne from 2020 to 2022. She was awarded the prestigious International Postgraduate Research Scholarship for doctoral studies in 2015 and received her PhD degree from The University of Western Australia in 2018. Her thesis ``Deep Learning for Action Recognition and Prediction'' has been awarded Dean's List-Honourable mention by The University of Western Australia in 2018. She was awarded the 1962 Medal for her work in video recognition technology by the Australian Computer Society in 2019, and APRS Early Career Researcher Award from Australian Pattern Recognition Society in 2020. Her main research topics include human action understanding and video analysis.
\end{IEEEbiography}

\end{document}